\documentclass{article} % For LaTeX2e
\usepackage{iclr2027_preprint,times}
\iclrfinalcopy
\usepackage{amsmath,amsfonts,bm}

\def\eqref#1{equation~\ref{#1}}
\def\1{\bm{1}}

\DeclareMathAlphabet{\mathsfit}{\encodingdefault}{\sfdefault}{m}{sl}
\SetMathAlphabet{\mathsfit}{bold}{\encodingdefault}{\sfdefault}{bx}{n}

\usepackage{hyperref}
\usepackage{xcolor}
\usepackage{url}
\usepackage{graphicx}
\usepackage{algorithm}
\usepackage{algpseudocode}
\usepackage{amsthm,booktabs,wrapfig,capt-of}

\title{Uncovering the Tells: Efficient Black-Box Audits for LLM Data Contamination}

\newcommand{\authcenter}[1]{\makebox[\dimexpr\textwidth-2\tabcolsep][c]{#1}}

\author{
Dominik B.~Lau$^{1*}$ \And
Filip Pawlicki$^{1,2*}$ \And
Jan Dubiński$^{1}$ \And
Sebastian Cygert$^{1,2}$ \AND
\authcenter{\mdseries$^{1}$NASK National Research Institute \quad\textbar\quad $^{2}$Gdańsk University of Technology} \\[4pt]
\authcenter{\texttt{\{firstname.lastname\}@nask.pl}} \\[4pt]
\authcenter{\small$^{*}$Equal contribution; order determined by a coin flip.}
}

\begin{document}

\maketitle

\begin{abstract}
    Existing methods for assessing training data contamination in Large Language Models achieve high accuracy, but fundamentally rely on gray-box access, rendering them inapplicable to fully black-box environments. This limitation forces auditors to rely on full-sequence sampling or rigid heuristics, both of which incur prohibitive inference costs and struggle to isolate the underlying memorization signal. To overcome this, we propose a generation-based black-box framework that infers context-induced information gain from targeted token discrepancies. We introduce a budget-constrained auditing regime, building on the empirical observation that the membership signal is highly sparse and concentrated in a few highly informative tokens. We formalize this sparsity, 
    showing why uniform sampling needs many more samples than a policy that targets informative positions.
    Empirically, our approach achieves state-of-the-art black-box separation accuracy while needing about $3\times$ fewer queries than uniform position sampling. Finally, we apply this methodology to closed frontier models, exploring its capacity and limits in surfacing potential training data overlaps on commercial APIs.
\end{abstract}

\section{Introduction}

Quantifying training data leakage in Large Language Models (LLMs) is critical for reliable model evaluation. As models increasingly saturate standard benchmarks \citep{phan2025hle, white2025livebench}, it remains unclear whether we measure true reasoning capabilities or merely the regurgitation of contaminated training data \citep{sainz2023nlp, magar2022datacontamination, yang2023rethinkingbenchmarkcontaminationlanguage, xu2024surveycontamination, fu2024loss, ravaut2025survey}. Furthermore, with growing scrutiny over data licensing and copyright compliance, the ability to transparently audit commercial models for memorized content is no longer just an academic pursuit, but a legal and operational necessity \citep{henderson2023fairuse}.

Existing contamination detectors predominantly operate under a \emph{gray-box} assumption, requiring access to per-token log-probabilities \citep{carlini2021extracting, shi2024detecting, xie2024recall, maini2024llmdatasetinferencedid, zhang2025mink, codec}. In reality, frontier models are served through APIs that expose little or no log-probability information \citep{casper2024blackbox, carlini2024stealing, finlayson2024logits}. To circumvent this, recent works have attempted \emph{black-box} approaches \citep{golchin2024time, tsguessing, duarte2024decop}. However, these methods are fundamentally behavioral: they rely on rigid prompting heuristics, such as forced multiple-choice formatting (A/B/C/D) \citep{duarte2024decop, tsguessing} or instruction-heavy wrappers \citep{golchin2024time}. These artificial constraints distort natural generation, obscuring the pre-training memorization signal they seek to detect.

To overcome this without log-probabilities, we propose a practical black-box audit framework. Our approach infers the underlying context-induced information gain \citep{xie2024recall, codec} by analyzing shifts in the model's greedy outputs. Specifically, we measure the discrepancy between the unconditioned token generation and the generation conditioned on a prefix from the evaluated dataset. By relying solely on the binary indicator of whether the target token is perfectly predicted, we construct a contamination score that captures parameter-level memorization.

\begin{figure}[t]
    \centering
    \includegraphics[width=1\linewidth]{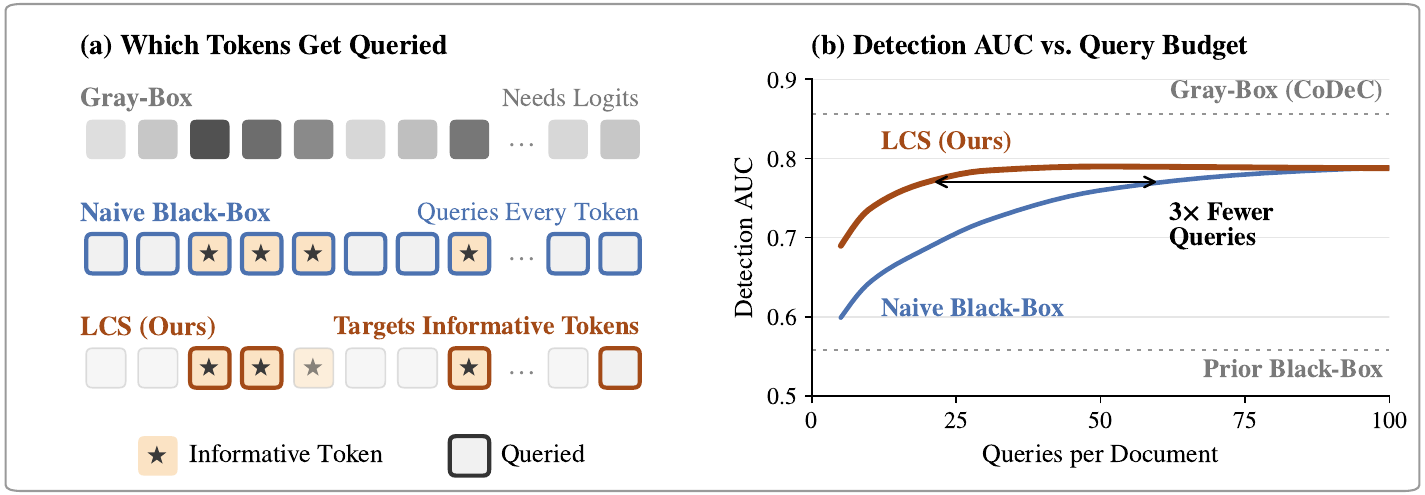}
    \caption{(a) With the same query budget, naive sampling mostly hits uninformative tokens, while LCS targets the informative ones. (b) Mean AUC vs.\ query budget on base models: LCS matches the naive black-box audit with $3\times$ fewer queries, far exceeds prior black-box methods, and approaches the gray-box CoDeC \citep{codec}. Details in Sections~\ref{main_results}--\ref{open-weighted-models}.}
    \label{fig:hero}
\end{figure}

Furthermore, a critical barrier to deploying such token-level audits on commercial endpoints is the prohibitive cost of scanning full sequences. To address this, we formalize a budget-constrained auditing regime optimized for real-world scenarios. Consistent with broader evidence that naively aggregated membership signal in LLMs is weak and diffuse \citep{duan2024membership}, we find that the membership signal is sparse: only about 10\% of positions change their top-1 prediction under context, concentrated within a handful of highly informative token positions \citep[cf.][]{shi2024detecting, zhang2025mink}, which we call ``tells''. As illustrated conceptually in Figure~\ref{fig:hero}, a naive full-sequence scan not only incurs significant API costs, but also aggregates uninformative noise, diluting the overall score. In contrast, targeting these sparse anomalies amplifies the true contamination signal. By employing a targeted token selection policy that identifies sequence anomalies based on string-matching heuristics \citep{hirschberg1975linear}, our method rapidly separates member and non-member datasets. This approach achieves state-of-the-art separation between seen and unseen data while reducing the required query budget several-fold.

In summary, our key contributions are:
\begin{itemize}
    \item \textbf{State-of-the-art black-box estimator and rigorous validation.} As introduced in Section~\ref{methodology}, we propose a contamination metric based on discrete context-induced information gain. We rigorously validate this framework across 19 open-weights checkpoints in Section~\ref{open-weighted-models}, demonstrating that our approach substantially outperforms existing generation-based black-box baselines and rivals gray-box alternatives such as zlib \citep{carlini2021extracting} without requiring logit access.
    
    \item \textbf{Statistical formulation of budget constraints.} We formalize the sample complexity penalty induced by signal sparsity, demonstrating why uniform sequence auditing is cost-prohibitive. Through the hyperparameter ablation detailed in Section~\ref{main_results}, we establish an auditing regime that saturates detection capability with a budget of just 16 tokens per sequence, drastically reducing API expenditure.
    
    \item \textbf{Audits of commercial frontier APIs.} To demonstrate the practical feasibility of independent audits, we deploy our budget-constrained method on production models in Section~\ref{api_audits}, with at most 3200 calls per evaluation. The results are mainly exploratory---we discuss their validity, showing where they are not reliable.
\end{itemize}

\section{Methodology}
\label{methodology}

We score the contamination of dataset $\mathcal{D}$ by its distributional similarity to model $\mathcal{M}$'s training corpus. Under strict black-box constraints \citep{casper2024blackbox}, we observe only single sampled output tokens $\hat x_t \sim \mathcal M(x_{<t})$, without access to output probabilities.

Our approach relies on context-induced information gain. For a document $x\in \mathcal D$, we prompt the model at position $t$ twice, with and without prepended context $C$, and record whether the context changes the top-1 prediction. Gray-box methods like CoDeC \citep{codec} score samples via continuous log-probability shifts: $\Delta_{\log}(x_t) = \log p(x_t \mid C, x_{<t}) - \log p(x_t \mid x_{<t})$. We instead use discrete indicator shifts $\delta_j \in \{-1, 0, 1\}$ over chosen positions $\mathcal{W} \subseteq \{1, \dots, T\}$:
\begin{equation}
    \Delta_{bb}(x, \mathcal{W}) = \sum_{j \in \mathcal{W}} \underbrace{\left( \boldsymbol{1} [\hat x_j = x_j \mid C, x_{<j}] - \boldsymbol{1} [\hat x_j = x_j \mid x_{<j}] \right)}_{\delta_j} = \sum_{j \in \mathcal{W}} \delta_j
\end{equation}

\subsection{Uncovering the ``Tells''}
\label{sec:tells}

This discrete formulation exploits the property that models gain little from conditioning on already-memorized data, whereas context sharply decreases entropy for unseen data \citep{codec} (Appendix~\ref{theoretical_foundations}). 

Crucially, this shift is mechanically coupled to unconditioned top-1 accuracy: a position can only be hurt ($\delta_j = -1$) if initially correct, and helped ($\delta_j = 1$) if initially incorrect. Our metric explicitly leverages this base-rate asymmetry. 
As Figure~\ref{fig:tells_visualization} illustrates, unseen documents (bottom) have lower baseline accuracy, creating greater scope for in-context learning \citep{brown2020language} to correct initially incorrect predictions (green).
Conversely, memorized documents (top) possess strong exact-match priors \citep{carlini2023quantifying}; introducing context actively interferes with these priors \citep{xie2024recall, codec}, systematically triggering the large available pool of negative shifts (red) and capturing the memorization signal without log-probabilities.

\begin{figure}[htbp]
  \centering
  \providecommand{\dahelp}[2]{\colorbox{green!#2!white}{\strut #1}}
  \providecommand{\dahurt}[2]{\colorbox{red!#2!white}{\strut #1}}
  \fbox{\parbox{0.98\linewidth}{
    \providecommand{\daidle}[1]{#1}
    \providecommand{\dadocsepskip}{3pt}
    \providecommand{\dadocsep}{\par\addvspace{\dadocsepskip}\noindent
      \hbox to\linewidth{\leaders\hbox to .7em{\hss.\hss}\hfil}%
      \par\addvspace{\dadocsepskip}\noindent\ignorespaces}

    \textbf{Seen Document:} \texttt{pile\_uspto\_backgrounds} \hfill \textit{Context hurts: 2 helped (green), 11 hurt (red)}
    \vspace{1pt}

    {\setlength{\fboxsep}{1pt}\sloppy\par\noindent
    \daidle{With} \daidle{a} \daidle{semiconductor} \daidle{testing} \daidle{system} \daidle{(}\daidle{including} \daidle{so}\daidle{-}\daidle{called} \daidle{an} \daidle{IC} \daidle{tester}\daidle{,} \daidle{L}\daidle{SI} \daidle{tester}\daidle{,} \daidle{and} \dahelp{so}{70} \dahurt{forth}{70}\daidle{;} \daidle{hereinafter} \daidle{referred} \daidle{to} \daidle{merely} \daidle{as} \dahurt{a}{70} \daidle{tester}\dahurt{),}{70} \daidle{it} \daidle{is} \daidle{necessary} \daidle{to} \daidle{supply} \daidle{a} \daidle{device} \daidle{as} \dahurt{a}{70} \daidle{test} \daidle{subject} \daidle{\{}\daidle{hereinafter} \daidle{referred} \daidle{to} \daidle{as} \daidle{a} \daidle{D}\daidle{UT} \dahurt{(}{70}\daidle{Device} \daidle{Under} \daidle{Test}\daidle{)\}} \dahelp{with}{70} \daidle{a} \daidle{voltage} \daidle{with} \daidle{high} \daidle{precision} \daidle{in} \daidle{order} \daidle{to} \daidle{inspect} \daidle{and} \daidle{test} \daidle{the} \dahurt{D}{70}\daidle{UT} \daidle{with} \daidle{good} \daidle{precision}\daidle{.} \daidle{Accordingly}\daidle{,} \dahurt{a}{70} \dahurt{tester}{70} \daidle{is} \daidle{provided} \daidle{with} \daidle{a} \daidle{power} \daidle{supply} \daidle{assembly} \daidle{capable} \daidle{of} \daidle{output}\daidle{ting} \daidle{a} \daidle{voltage} \dahurt{with}{70} \daidle{high} \daidle{precision}\daidle{.} \daidle{Determination} \daidle{on} \daidle{whether} \dahurt{or}{70} \daidle{not} \daidle{a} \daidle{D}\daidle{UT} \daidle{is} \daidle{acceptable} \daidle{has} \dahurt{been}{70} \daidle{made} \daidle{by} \daidle{taking} \daidle{measurements} \daidle{on}\par}

    \dadocsep
    \textbf{Unseen Document:} \texttt{patent\_2025} \hfill \textit{Context helps: 10 helped (green), 1 hurt (red)}
    \vspace{1pt}

    {\setlength{\fboxsep}{1pt}\sloppy\par\noindent
    \daidle{1}\daidle{.} \dahelp{Field}{70} \daidle{of} \daidle{the} \dahelp{Dis}{70}\daidle{closure} \daidle{The} \dahurt{present}{70} \dahelp{disclosure}{70} \daidle{generally} \daidle{relates} \daidle{to} \dahelp{an}{70} \daidle{electronic} \daidle{device} \daidle{and} \dahelp{a}{70} \daidle{method} \daidle{of} \daidle{manufacturing} \daidle{an} \daidle{electronic} \daidle{device}\daidle{.} \daidle{2}\daidle{.} \daidle{Description} \daidle{of} \daidle{the} \daidle{Related} \daidle{Art} \daidle{An}\daidle{ten}\daidle{na} \daidle{package}\daidle{(}\daidle{s}\daidle{)} \daidle{having} \daidle{antennas} \daidle{for} \daidle{signal} \daidle{(}\daidle{e}\daidle{.}\daidle{g}\daidle{.,} \daidle{radio} \daidle{frequency} \daidle{(}\daidle{RF}\daidle{)} \daidle{signal}\daidle{)} \dahelp{transmission}{70} \daidle{and} \daidle{reception} \daidle{may} \daidle{include} \daidle{an} \daidle{antenna} \daidle{layer} \dahelp{and}{70} \daidle{an} \daidle{RF} \daidle{routing} \daidle{layer} \daidle{electrically} \daidle{connected} \daidle{thereto}\daidle{.} \daidle{In} \daidle{some} \daidle{existing} \daidle{techniques} \daidle{for} \daidle{forming} \dahelp{an}{70} \daidle{electromagnetic} \daidle{interference} \daidle{(}\daidle{EM}\daidle{I}\daidle{)} \daidle{shielding} \daidle{layer} \daidle{on} \dahelp{an}{70} \daidle{antenna} \daidle{package}\daidle{,} \dahelp{a}{70} \daidle{tape} \daidle{(}\daidle{e}\daidle{.}\daidle{g}\daidle{.,} \daidle{a} \daidle{thermally} \daidle{stable} \daidle{tape}\daidle{)} \daidle{may} \daidle{be} \daidle{utilized} \daidle{to} \daidle{protect} \daidle{the} \daidle{antenna}\par}
  }}
  \caption{Positions where adding context ($K=3$) flips the top-1 prediction (evaluated on Pythia-410M). {\setlength{\fboxsep}{1pt}\protect\dahelp{Green}{70}} highlights ($\delta_j = 1$) indicate shifts toward the ground-truth token, while {\setlength{\fboxsep}{1pt}\protect\dahurt{red}{70}} highlights ($\delta_j = -1$) indicate shifts away. Crucially, for unseen text there are significantly more tokens where the context helps (green), while for seen text the context predominantly hurts (red) by interfering with exact memorization.}
  \label{fig:tells_visualization}
  \vspace{-10pt}
\end{figure}
\subsection{Budgeted Auditing and Linear Amplification}
\label{sec:budget}

Auditing all $T$ positions requires $2 \cdot |\mathcal{D}| \cdot T$ queries, which is cost-prohibitive. We reduce this by evaluating a subset $\mathcal D_n \subseteq \mathcal D$ ($|\mathcal{D}_n| = n$) and limiting queries to $W$ positions via a selection policy $\pi$. 

Targeting the budget $W$ exclusively on relevant tokens ($\delta_j \neq 0$) accelerates score convergence by $T / T_{\text{rel}}$. We formalize this via Bernstein's inequality \citep{boucheron2013concentration}. Let $T_{\text{rel}}$ denote the relevant token count, defining a sparsity ratio $\rho = T_{\text{rel}}/T$. Estimating the conditional mean signal $\mu = \mathbb{E}[\delta_j \mid \delta_j \neq 0]$ (which determines the score $\mathcal{S} = (1-\mu)/2$), uniformly sampling $W$ tokens gives expected shift $\mathbb{E}[Y_i] = \rho \mu$ and variance $\operatorname{Var}(Y_i) \le \rho$. For an unbiased empirical mean $\hat{\mu}_{\text{naive}} = \frac{1}{\rho W} \sum_{i=1}^W Y_i$, the deviation $\epsilon \in (0,1)$ is bounded by:
\begin{equation}
    \mathbb{P}\left(|\hat{\mu}_{\text{naive}} - \mu| \ge \epsilon\right) = \mathbb{P}\left( \left| \frac{1}{W} \sum_{i=1}^W Y_i - \rho \mu \right| \ge \rho \epsilon \right) \le 2 \exp\left( - \frac{W \epsilon^2 \rho}{2(1 + 2\epsilon/3)} \right)
\end{equation}

Conversely, if an ideal targeted policy strictly selects relevant tokens, the expected shift is $\mathbb{E}[Z_i] = \mu$. The empirical mean $\hat{\mu}_{\text{target}} = \frac{1}{W} \sum_{i=1}^W Z_i$ bounds the error without the sparsity penalty:
\begin{equation}
    \mathbb{P}\left(|\hat{\mu}_{\text{target}} - \mu| \ge \epsilon\right) \le 2 \exp\left( - \frac{W \epsilon^2}{2(1 + 2\epsilon/3)} \right)
\end{equation}

To achieve identical statistical error bounds, the required budgets must satisfy $W_{\text{naive}}\, \rho = W_{\text{target}}$. Substituting $\rho = T_{\text{rel}}/T$, we arrive at the sample complexity:
\begin{equation}
    W_{\text{naive}} = \frac{1}{\rho} W_{\text{target}} = \frac{T}{T_{\text{rel}}}\, W_{\text{target}}
\end{equation}
This rate is tight, since $\operatorname{Var}(\hat{\mu}_{\text{naive}}) = (1/\rho - \mu^2)/W \ge \operatorname{Var}(\hat{\mu}_{\text{target}})/\rho$.

Because the membership signal is sparse ($T_{\text{rel}} \ll T$), blind
full-sequence auditing is inefficient: to match the variance of an Oracle that
queries only informative positions, uniform sampling needs about $1/\rho$ times
as many positions. This motivates \emph{targeted} selection, i.e., spending the
query budget on positions likely to be informative.
\subsection{Token and Context Selection Policies}
We compare policies $\pi(x, W)$ that select $W$ positions of a document $x$:
\begin{itemize}
        \item \textbf{Uniform / Prefix / Suffix}: Sample $W$ positions uniformly \citep[cf.][]{xu2024benchmarking}, or select the first/last $W$ positions. While Prefix often captures early shifts, Uniform and Suffix waste budget on uninformative positions ($\delta_j = 0$), incurring the full $\mathcal O(1/\rho)$ sparsity penalty.
        \item \textbf{LCS (Ours)}: Select $W$ tokens within the Longest Common Subsequence (LCS) \citep{hirschberg1975linear} between target $x$ and context $C$. LCS offers two benefits over Uniform. First, a gain in density: a higher fraction of its selected positions have $\delta_j \neq 0$.  Second, it targets different positions: LCS over-represents positions where the context helps ($\delta_j=+1$), and the sign balance $\mathcal{S}$ of the shifts at these positions differs clearly between seen and unseen data. We further explain the details of this heuristic and explore the traits in Appendix \ref{lcs_details}.
        \item \textbf{Oracle}: Selects strictly known $\delta_j \ne 0$ positions (idealized reference).
\end{itemize}

\paragraph{Context Selection ($\pi_c$).} We construct context $C$ by ranking remaining documents in $\mathcal{D}$ via Jaccard similarity \citep{jaccard1912distribution} to $x$, extracting $\tau=100$ tokens from the top $K=3$ matches \citep[cf.][]{liu2022makes} (ablated in Appendix~\ref{context_policy}).
\newpage
\section{Budget-Accuracy Trade-Off}
\label{main_results}

\paragraph{Experimental Setup.}
We evaluate the framework on 11 pre-trained off-the-shelf models (detailed in Section~\ref{open-weighted-models}), relying strictly on base checkpoints to isolate pre-training memorization from fine-tuning variance. We compare the black-box heuristics (\textit{Prefix, Suffix, Uniform, LCS}) against the \textit{Oracle}, an idealized reference rather than a practical policy, since identifying $\delta_j \ne 0$ a priori requires querying every position.
 
\begin{figure}[htbp]
    \centering
    \includegraphics[width=0.49\linewidth]{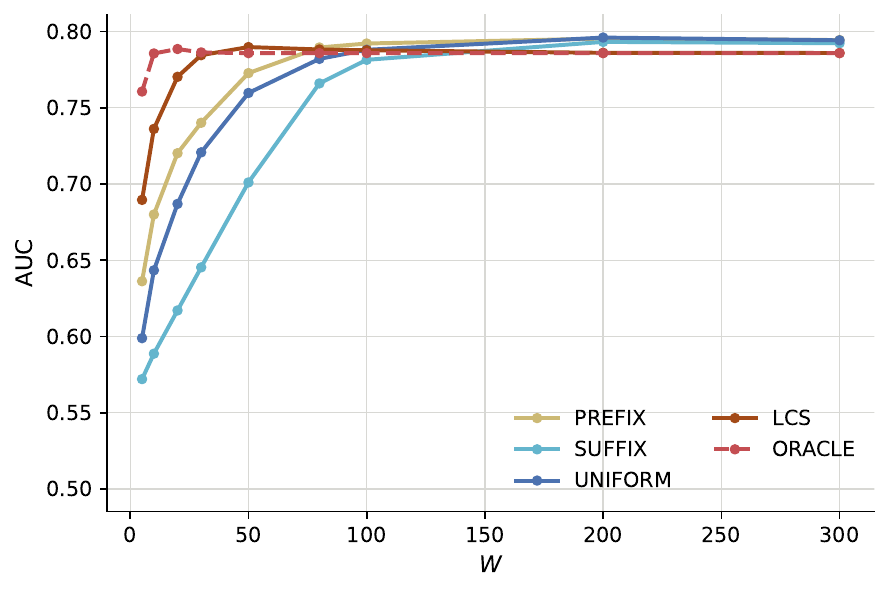}
    \includegraphics[width=0.49\linewidth]{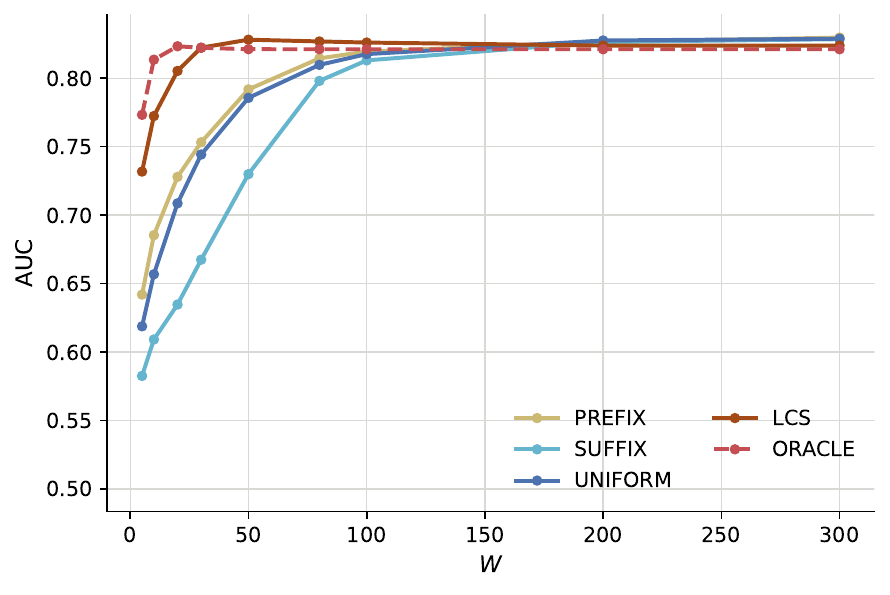}
    \caption{Cost-effectiveness of token selection policies across 15 models (11 pre-trained base models on the left and 4 post-trained Pythia \citep{biderman2023pythia} checkpoints on the right). Pre-trained (left) and post-trained (right) results show mean document-level AUC per budget. LCS demonstrates superior sample efficiency, reaching the level close to the Oracle at $W \le 30$.}
    \label{fig:window_budget_tradeoff}
    % \vspace{em}
\end{figure}

\paragraph{Convergence of Token Selection Policies.}
Figure~\ref{fig:window_budget_tradeoff} is consistent with our theoretical framework. The \textit{Oracle} policy rapidly achieves optimal separation, as predicted in Section~\ref{sec:budget}: only $\rho \approx 0.1$ of positions are relevant, implying a $1/\rho \approx 9.5\times$ budget reduction over \textit{Uniform}, and we observe at least $8.8\times$ at AUC~0.75. Among black-box heuristics, \textit{Uniform} and \textit{Suffix} converge slowest, indicating later tokens yield minimal information gain. Conversely, \textit{Prefix} captures relevant positions faster due to higher marginal entropy of early tokens \citep{Olsson2022}. Crucially, \textit{LCS} is the most sample-efficient heuristic, saturating detection at $W \le 30$ and is close to the Oracle. Naive policies catch up only at near-exhaustive budgets ($W \ge 100$), which negate the premise of a budgeted audit.
 
\begin{wrapfigure}{r}{0.48\textwidth}
    \centering
    \vspace{-15pt}
    \includegraphics[width=\linewidth]{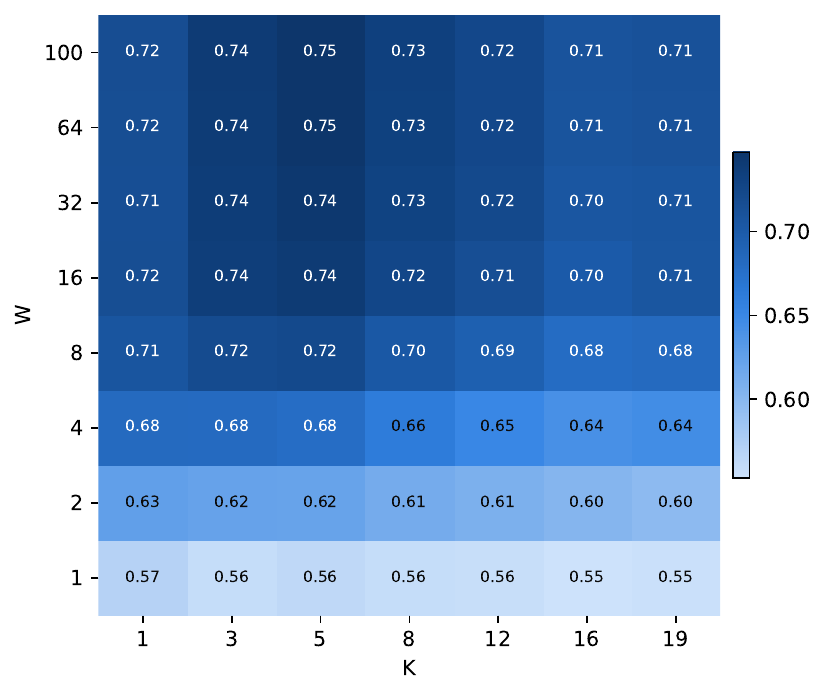}
    \vspace{-20pt}
    \caption{Hyperparameter ablation grid across 7 off-the-shelf models, displaying the document-level AUC given varying token budgets ($W$) and number of conext document ($K$).}
    \label{fig:ablation_grid}
\end{wrapfigure}
 
\textbf{Hyperparameter Ablation.}
To optimize large-scale API audits, we grid-search over the number of context documents $K$ and the token budget $W$. Figure~\ref{fig:ablation_grid} maps the mean AUC across the 7 ablation models. Performance exhibits a steep initial gain that plateaus rapidly. Utilizing $K=3$ documents and a strictly constrained budget of $W=16$ tokens achieves an AUC of 0.74. Expanding to $K=5$ documents and $W \in \{64, 100\}$ tokens yields only a marginal improvement to 0.75. Consistent with Appendix~\ref{context_policy}, the signal depends far more on \emph{which} context documents are selected than on \emph{how many}, so $K$ acts primarily as a cost parameter. We therefore adopt $K=3$ and $W=16$ as a cost-effective configuration. This plateau is consistent with the sparsity motivation of Section~\ref{sec:budget}: for LCS-selected positions, beyond 16 tokens the additional positions add little.

\section{Auditing Public Models}
\label{open-weighted-models}

\paragraph{Comparison with Black-Box Methods.} Following the protocol of \citet{codec}, we evaluate on 11 open-weights base checkpoints with known training data \citep{black2021gptneo, biderman2023pythia, peng2023rwkv, groeneveld2024olmo}, spanning the Pile \citep{gao2020pile} and Dolma \citep{soldaini2024dolma} families. Our unseen dataset pools strictly post-date each family's pre-training cutoff (Appendix~\ref{ots_setup}); since temporal splits can introduce distribution shifts unrelated to membership \citep{duan2024membership, das2025blind, meeus2025sok}, we discuss this limitation in Section~\ref{sec:discussion} and investigate it in Appendix~\ref{apd:blind_baselines}.

We compare LCS against generation-based black-box baselines: verbatim extraction \citep{carlini2021extracting, carlini2023quantifying}, SA-MIA \citep{kaneko2024samplingbasedpseudolikelihoodmembershipinference}, and SimMIA \citep{yi2026simmia}, the latter two under their original configurations. Since the original extraction attack requires log-probabilities, we use a graded black-box variant of the protocol of \citet{carlini2023quantifying}: from a 50-token prefix, we greedily decode 50 tokens and score the normalized length of their longest prefix matching the ground truth. We audit 100 documents per dataset with $W=16$ and $K=3$ (Section~\ref{main_results}).

\begin{figure}[htbp]
    \centering
    \includegraphics[width=1\linewidth]{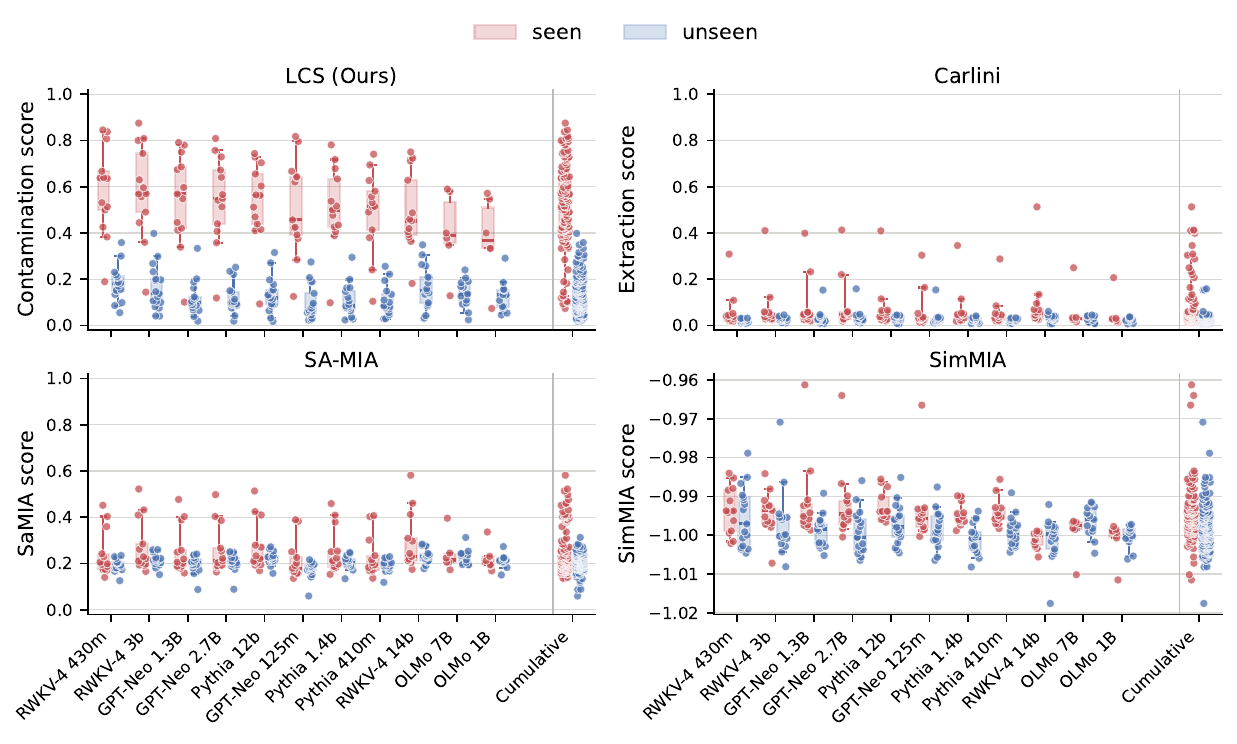}
    
    \caption{Contamination scores across 11 base checkpoints. Red and blue denote seen and unseen corpora, respectively. Box plots show quartiles and 5–95\% whiskers with dots as individual corpora. The 2x2 grid compares our approach against standard black-box baselines. \textbf{Top-left:} Our LCS method achieves clear separation between seen and unseen data (mean dataset-level AUC $0.95$). \textbf{Remaining panels:} SA-MIA is near chance ($0.56$), Carlini ($0.73$) and SimMIA ($0.67$) separate the datasets only partially.}
    \label{fig:separations}
\end{figure}

As detailed in Table~\ref{tab:table_ots_auc_blackbox_full} and Figure~\ref{fig:separations}, LCS vastly outperforms all baselines (mean dataset-level AUC 0.952 vs.\ 0.562--0.732). Standard IID splits (e.g., train vs. validation) are ill-suited for corpus-level contamination, as members and non-members share substantial n-gram overlap, blurring generalization and memorization \citep{duan2024membership}. Against domain-related, temporally separated controls, separation is striking at the document level: \texttt{pile\_stackexchange} vs. \texttt{gsm8k\_2021} \citep{cobbe2021gsm8k} (AUC 0.930) and \texttt{math500\_2021} \citep{hendrycks2021math, lightman2024verify} (AUC 0.761), or \texttt{pile\_github} vs. \texttt{humaneval\_2021} \citep{chen2021humaneval} (AUC 0.896). Mechanistically, domain shift alone is difficult to reconcile with helpful semantic context actively \emph{hurting} predictions on seen datasets ($\delta_j = -1$); this interference is instead consistent with the disruption of exact-match verbatim priors \citep{xie2024recall, codec}.

\begin{wraptable}{r}{0.6\textwidth}
\vspace{-1em}
\centering
\caption{Black-box baseline comparison, per base model. Dataset-level AUC separating each model's seen corpora from its unseen ones (one score per dataset).}
\label{tab:table_ots_auc_blackbox_full}
\small
\resizebox{\linewidth}{!}{%
\begin{tabular}{@{}lcccc@{}}
\toprule
Model & Carlini & SA-MIA & SimMIA & LCS (Ours) \\
\midrule
GPT-Neo 125M & 0.726 & 0.558 & 0.729 & \textbf{0.977} \\
GPT-Neo 1.3B & 0.776 & 0.596 & 0.801 & \textbf{0.964} \\
GPT-Neo 2.7B & 0.764 & 0.572 & 0.787 & \textbf{0.977} \\
Pythia 410M & 0.798 & 0.582 & 0.860 & \textbf{0.968} \\
Pythia 1.4B & 0.779 & 0.601 & 0.882 & \textbf{0.968} \\
Pythia 12B & 0.740 & 0.548 & 0.783 & \textbf{0.950} \\
RWKV-4 430M & 0.769 & 0.596 & 0.593 & \textbf{0.973} \\
RWKV-4 3B & 0.827 & 0.591 & 0.710 & \textbf{0.959} \\
RWKV-4 14B & 0.808 & 0.591 & 0.403 & \textbf{0.968} \\
OLMo 1B & 0.528 & 0.472 & 0.526 & \textbf{0.859} \\
OLMo 7B & 0.542 & 0.472 & 0.346 & \textbf{0.910} \\
\midrule
\textit{Mean} & 0.732 & 0.562 & 0.674 & \textbf{0.952} \\
\bottomrule
\end{tabular}%
}
\vspace{-1em}
\end{wraptable}

Conversely, SA-MIA collapses to near-chance performance (AUC $\approx 0.56$), with seen and unseen distributions overlapping almost completely (Figure~\ref{fig:separations}), while Carlini's attack (0.732) and SimMIA (0.674) improve moderately, but far below LCS (0.952). We attribute this weakness to their reliance on (near-)verbatim reproduction during free-running generation—an excessively high bar for single-epoch pre-training prose \citep{duan2024membership}. These baselines only registered elevated scores on templated, low-entropy corpora (e.g., \texttt{pile\_github}), consistent with extraction being dominated by duplicated sequences \citep{kandpal2022deduplicating, carlini2023quantifying}.

\begin{wraptable}{r}{0.5\textwidth} 
\centering
\vspace{-2em}
\caption{Comparison of our black-box method (LCS) to competing gray-box methods requiring log-probabilities (dataset-level metrics).}
\label{tab:table_ots_auc}
\small
\begin{tabular*}{\linewidth}{@{\extracolsep{\fill}}lcc} 
\toprule
Method & AUC & TPR@5\%FPR \\
\midrule
Loss & 0.585 & 0.120 \\
Min-K\% & 0.599 & 0.134 \\
zlib & 0.759 & 0.147 \\
CoDeC & \textbf{0.961} & 0.698 \\
LCS (Ours)  & 0.952 & \textbf{0.893} \\
\bottomrule
\end{tabular*}
\vspace{-10pt}
\end{wraptable}

Unlike instance-level MIAs \citep{shokri2017}, which perform near chance on real pre-training data due to near single-epoch training and fuzzy membership boundaries \citep{duan2024membership}, dataset-level approaches aggregate weak per-instance signals \citep{maini2024llmdatasetinferencedid, puerto2025scaling, codec}. Our approach extends this idea to strictly black-box settings by aggregating targeted, context-induced prediction shifts across an entire corpus, without requiring unprompted verbatim reproduction.

\begin{wraptable}{r}{0.5\textwidth}
\vspace{-2em}
\centering
\caption{Dataset-level detection performance on post-trained checkpoints. The contamination signal is preserved across post-training.}
\label{tab:post_train}
\small
\setlength{\tabcolsep}{4pt}
\begin{tabular*}{\linewidth}{@{\extracolsep{\fill}}lcc} 
\toprule
Model & AUC & TPR@5\%FPR \\
\midrule
Pythia 410M & 0.968 & 0.846 \\
\quad + SFT & 0.968 & 0.923 \\
\quad + DPO & 0.968 & 0.923 \\
\midrule
Pythia 1.4B & 0.968 & 0.923 \\
\quad + SFT & 0.968 & 0.923 \\
\quad + DPO & 0.968 & 0.923 \\
\midrule
RWKV-4 3B & 0.959 & 0.846 \\
\quad + SFT & 0.964 & 0.923 \\
\midrule
RWKV-4 14B & 0.968 & 0.923 \\
\quad + SFT & 0.982 & 0.923 \\
\midrule
OLMo 7B & 0.910 & 0.833 \\
\quad + SFT & 0.859 & 0.833 \\
\quad + Instruct & 0.859 & 0.833 \\
\bottomrule
\end{tabular*}
\vspace{-2em}
\end{wraptable}

\noindent\textbf{Comparison with Gray-Box Methods.} Despite the query constraint ($W=16$), our black-box LCS policy (dataset-level AUC 0.952; Table~\ref{tab:table_ots_auc}) outperforms Loss \citep{yeom2018privacy} (0.585), Min-K\% \citep{shi2024detecting} (0.599) and zlib \citep{carlini2021extracting} (0.759), and is on par with the gray-box CoDeC \citep{codec} (0.961), with a higher TPR@5\%FPR (0.893 vs.\ 0.698).
Uncalibrated likelihood scores such as Loss and Min-K\% conflate memorization with the intrinsic predictability of text, which zlib partially corrects by normalizing against compression entropy \citep{carlini2021extracting}; our metric may play an analogous calibrating role, as each position is compared against its own unconditioned prediction. 

\noindent\textbf{Audit of Post-Trained Models.} We evaluate aligned checkpoints \citep{peng2023rwkv, groeneveld2024olmo} to test whether post-training obscures the membership signal (Table~\ref{tab:post_train}). While SFT and DPO \citep{ouyang2022training, rafailov2023dpo} suppress unprompted regurgitation \citep{nasr2025scalable}, our ground-truth prefixes avoid free-running rollouts altogether. Dataset-level AUC is unchanged or higher after post-training for all models except OLMo 7B (0.910 to 0.859 after SFT), suggesting that post-training largely does not erase the pre-training membership signal \citep{nasr2025scalable, lin2024unlocking, qi2025safety}.
\section{Auditing Frontier Models}
\label{api_audits}

To demonstrate the practical feasibility of our framework, we deploy it against frontier models via public inference APIs under strict black-box constraints. Lacking public ground truth for the training mixtures of these commercial systems, this audit serves as an exploratory proof-of-concept rather than a definitive claim of contamination \citep{zhang2025position}. To query near-deterministic next-token predictions, we dispatch requests at temperature 0 (0.6 for Kimi-k2.6, due to provider constraints), allocate token positions via LCS, and exclusively use endpoints supporting assistant prefill to isolate the memorization signal from instruction bias. 
Prefill preserves plain continuation: the target token is the top-1 prediction at 72.6\% of positions with prefill versus 74.1\% without any template (Appendix~\ref{modalify_shift}), so the API audits probe the same next-token behavior as our open-weights experiments.

We evaluate Kimi-k2.6 \citep{kimi2025k2, kimi2026k26}, DeepSeek-V3 \citep{deepseekai2024v3}, DeepSeek-V4-Pro \citep{deepseekai2026v4}, Claude Sonnet 4.5 \citep{anthropic2025sonnet45}, and Claude Opus 4.5 \citep{anthropic2025opus45}. Allocating a budget of $K = 3$ context documents and $W = 16$ tokens for $n=100$ sequences yields a highly efficient upper bound of $B \le 3{,}200$ API calls per evaluation. A detailed breakdown is provided in Appendix~\ref{api_logistics}.

\begin{figure}[htbp]
    \centering
    \vspace{-0.2cm}
    \includegraphics[width=1.0\linewidth]{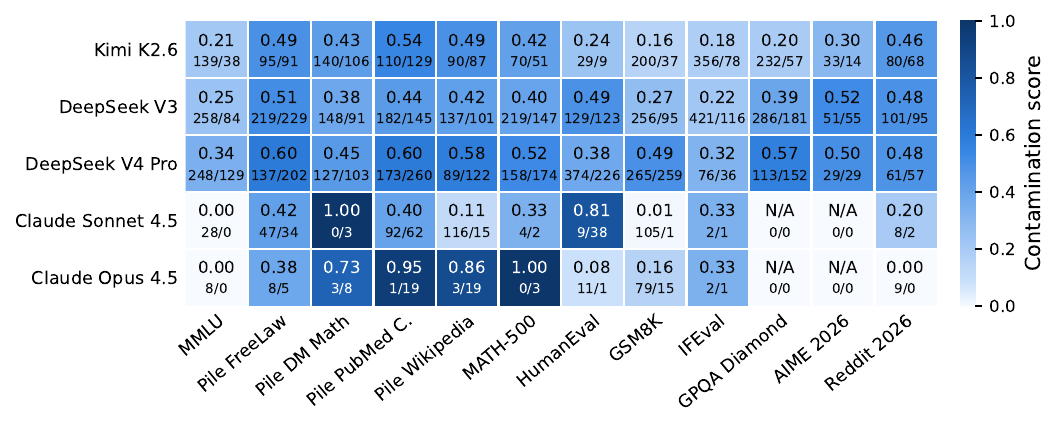}
    \vspace{-0.3cm}
    \caption{Contamination scores ($\mathcal{S}$) for frontier models evaluated via public APIs. Recent datasets (\texttt{aime2026}, \texttt{reddit\_2026}) act as internal negative controls to account for varying base capabilities. Below each score we present counts of positive / negative flips.}
    \vspace{-0.1cm}
    \label{fig:frontier_heatmap}
\end{figure}

As base capabilities and alignment strictness vary across models, we replace a universal threshold with a per-model baseline built from post-2025 datasets. These controls are smaller (e.g., $n=25$ for \texttt{aime2026}), so the baseline exhibits higher variance. Moreover, since the training cutoffs of the most recent models (Kimi-k2.6, DeepSeek-V4-Pro) are not publicly specified, we cannot rule out partial exposure to these controls, which would bias their baselines upward and make our estimates conservative.

Figure~\ref{fig:frontier_heatmap} reveals the difficulty of auditing frontier models. For Claude Opus~4.5 and Sonnet~4.5 the audit is not reliable (despite using assistant prefill, 93\% of no-context replies are in assistant voice and the target token appears at $\approx 1\%$ of positions, see Appendix~\ref{reliability}). Kimi-k2.6 (which we could only query at non-zero temperature) and both DeepSeek models do not separate clearly, with expected positives staying close to their 2026 controls. MMLU \citep{hendrycks2021mmlu} scores at or below the controls on all five models.
We attribute the weak signal on saturated text to the base-rate mechanics of Section~\ref{sec:tells}. As frontier models approach near-perfect unconditioned accuracy on ubiquitous prose or structured benchmarks such as MMLU \citep[cf.][]{carlini2023quantifying}, the pool of context-induced flips ($\delta_j \ne 0$) shrinks, and the score of such a dataset rests on few relevant positions. Argmax-based audits therefore lose sensitivity as model capabilities saturate, motivating dynamic difficulty scaling in future audits \citep{zhu2024dyval, white2025livebench}.

\section{Discussion}
\label{sec:discussion}

Observing only discrete output changes under small query budgets suffices to rival several full-access methods, lowering the barrier to third-party verification of copyright compliance and benchmark integrity on commercial APIs \citep{henderson2023fairuse, sainz2023nlp}. However, black-box evidence alone cannot prove that specific data was used in training \citep{casper2024blackbox, zhang2025position}.

\paragraph{Limitations and Future Work.}
Our framework faces four main constraints. First, it relies on assistant prefill, which some providers have already withdrawn for newer models, motivating adaptations to strictly conversational APIs. Second, deliberate provider obfuscation (e.g., output perturbations) could disrupt the metric \citep{carlini2024stealing}. Third, our negative controls rely on temporally separated corpora, which may introduce distribution shifts unrelated to membership \citep{duan2024membership, das2025blind, meeus2025sok}. Finally, the score depends on baseline accuracy: on low-entropy text, or as frontier models grow more capable, unprompted predictions align closer with ground truth, compressing the membership signal \citep[cf.][]{carlini2023quantifying}. Future research should explore dynamic difficulty scaling \citep{zhu2024dyval, white2025livebench}, e.g., targeting complex or low-frequency text regions \citep[cf.][]{shi2024detecting, zhang2025mink}.

\section{Related Work}

\paragraph{In-Context Conditioning and Memorization.}
LLMs memorize training data \citep{tirumala2022memorization, carlini2023quantifying}, while in-context copying mechanisms such as induction heads \citep{Olsson2022} let the prompt override these memorized priors. Context-induced shifts can thus quantify contamination \citep{xie2024recall, codec, yi2026simmia}; we formalize this as information gain (Appendix~\ref{theoretical_foundations}) and show that a discrete indicator retains most of this signal.

\paragraph{Gray-Box Data Contamination.}
Gray-box detectors score model likelihoods \citep{yeom2018privacy} with zlib normalization \citep{carlini2021extracting}, focus on low-probability tail tokens \citep{shi2024detecting, zhang2025mink}, or calibrate via perturbations \citep{mattern-etal-2023-membership, mitchell2023detectgptzeroshotmachinegeneratedtext}, token frequencies \citep{zhang2024pdd}, or conditional prefixes \citep{xie2024recall}. Others test benchmark-order exchangeability \citep{oren2024proving}, learn detectors over output distributions \citep{barshalom2025tokenprobabilitieslearnablefast}, or aggregate instance scores into dataset-level tests \citep{maini2024llmdatasetinferencedid, puerto2025scaling}. All require log-probabilities, which major providers increasingly restrict or disable \citep{carlini2024stealing, finlayson2024logits}.

\paragraph{Black-Box Contamination Detection.}
For sampling-only APIs \citep{casper2024blackbox}, instruction-dependent methods force structured tasks \citep{golchin2024time, tsguessing, duarte2024decop}, conflating instruction failure with non-memorization. Generation-based alternatives rely on verbatim extraction \citep{carlini2021extracting, nasr2025scalable}, n-gram prediction accuracy \citep{xu2024benchmarking}, output-distribution peakedness \citep{dong2024generalization}, or pseudo-likelihoods \citep{kaneko2024samplingbasedpseudolikelihoodmembershipinference, yi2026simmia}. These require either repeated free-running generations, prone to error propagation \citep{bengio2015scheduled}, or untargeted position sampling, incurring high query costs. We instead combine targeted token selection with teacher forcing \citep{williams1989learning} to isolate informative ``tells.''

\section{Conclusion}

In this work, we bridge the gap between gray-box contamination metrics and restrictive commercial APIs. By targeting the sparse membership signal with heuristics like LCS, we drastically reduce query budgets relative to full-sequence black-box audits, while our contamination score rivals several full-access methods. This provides a practical foundation for rigorous, independent audits of models in the wild.

\clearpage
\subsection*{AI Use Statement}

In this work, we used generative AI tools to draft sections of the paper, aid and polish the writing of the manuscript, and as a coding assistant to help implement and debug our experimental pipeline. We have not used generative AI tools to formulate the core scientific claims or to generate the empirical results presented in this study.

We have reviewed all AI-assisted work. Specifically, all AI-generated code was systematically verified and tested for correctness by the authors to ensure accurate execution of the API evaluations and metric calculations. Additionally, all AI-drafted and polished text was rigorously reviewed, edited, and validated by the authors to guarantee scientific accuracy. We take responsibility for the final content of this work, including text, claims or artifacts produced with the aid of generative AI.

\subsection*{Ethics Statement}

This research aims to promote transparency and accountability in Large Language Models (LLMs) by enabling independent, budget-constrained audits for data contamination. While Membership Inference Attacks can raise data privacy concerns, our methodology is explicitly designed for corpus-level detection rather than instance-level extraction. Crucially, the framework requires the auditor to already possess the target document to construct the conditional prefix, meaning it cannot be used by malicious actors to blindly extract novel Personally Identifiable Information (PII) or sensitive intellectual property.

Additionally, all evaluations on commercial models were conducted via public APIs in strict compliance with standard usage terms. Through mathematical optimization, our approach drastically minimizes the compute overhead and energy consumption typically associated with generation-based audits, ensuring that large-scale transparency research can be performed sustainably.

\subsection*{Reproducibility Statement}

To ensure full reproducibility of our results, we have made the complete source code publicly available at \url{https://github.com/NASK-AISafety/uncovering-the-tells}. This repository includes the reference implementation of our black-box estimator as well as a standalone tool for conducting custom dataset audits. All experimental setups, including data partitioning, model checkpoints, and LoRA fine-tuning hyperparameters, are exhaustively documented in Appendix~\ref{ots_setup} and \ref{app:finetuning_noise}. Finally, the exact prompt structures and provider-specific configurations necessary to replicate the frontier model API evaluations are outlined in Appendix~\ref{api_logistics}.

\clearpage

\bibliography{main}
\bibliographystyle{iclr2027_conference}

\appendix

\newpage

\section{Theoretical Foundations}\label{theoretical_foundations}
 
Let $\theta$ denote the parameters of the model that has seen dataset $\mathcal D$ during training and $\phi$ denote the uncontaminated parameters. An information-theoretic perspective on various data contamination/data inference scoring among many different types of attacks \citep{carlini2021extracting, mitchell2023detectgptzeroshotmachinegeneratedtext, mattern-etal-2023-membership, xie2024recall, codec} reveals the underlying use of the difference of conditioned and marginal distribution entropy shift i.e.
\begin{equation}
    H(x) - H(x\mid C) = \operatorname{IG}(x, C) > 0
\end{equation}
thus, the effect is amplified with respect to what context $C$ was used and how strongly it narrows down the entropy of the model distribution. When a model has seen $\mathcal{D}$ during training, much of $C$ is already encoded in $\theta$, therefore, $\operatorname{IG}_\theta(x,C) \rightarrow 0$ \citep{codec}. Conversely, $\operatorname{IG}_\phi(x,C) \gg 0$ as $C \subseteq \mathcal{D}$ is a highly-informative descriptor of the audited dataset. Naturally, $\Delta_{\log} \approx \operatorname{IG}(x, C)$. $\Delta_{bb}$ is a projection of the logprob shift onto a comparison of two Kronecker deltas, equivalently approximating, albeit more crudely, the context-induced information gain.
 
\paragraph{Remark on Teacher Forcing.}
The statistical guarantees provided in Section~\ref{sec:budget} rely on the assumption that the sampled indicator shifts $Y_i$ are drawn independently. While standard autoregressive modeling violates this, our framework operates strictly in a teacher-forcing regime \citep{williams1989learning}. For every evaluated position $j$, the prompt is constructed using the true ground-truth prefix $x_{<j}$ from the target dataset, rather than the model's own prior generations. Because of this, errors cannot propagate between positions \citep{bengio2015scheduled}, and each $\delta_j$ is a fixed function of the document that does not depend on which other positions are queried; positions drawn at random therefore yield independent shifts $Y_i$, satisfying the independence assumption \citep[see][Section~6]{hoeffding1963probability}.
 
\paragraph{Autocorrelation of the True Signal.}
While the sampling process is independent due to teacher forcing, the underlying true membership signal within the text is strongly autocorrelated. Memorization occurs in continuous chunks rather than uniformly distributed single tokens \citep{carlini2023quantifying, nasr2025scalable}. Uniform sampling fails to exploit this structure, thereby paying the full $\mathcal O(\frac{1}{\rho})$ sparsity penalty derived in our budget allocation proof. Conversely, policies like LCS explicitly exploit this autocorrelation by targeting dense regions of lexical overlap, bridging the gap to the theoretical optimum, although positions within one window are then correlated, so the guarantees of Section~\ref{sec:budget} apply to them only approximately.

\section{Detailed Experimental Setup}
\label{ots_setup}

\subsection{Public Models}
\label{setup_models}
We evaluate 11 base checkpoints spanning diverse parameter scales (125M to 14B) whose pre-training corpora (The Pile and Dolma) \citep{gao2020pile, soldaini2024dolma} are publicly accessible, so the membership of the seen corpora is known. Table \ref{tab:models} lists the base models and post-trained derivatives, including RWKV-4 as a non-transformer control architecture.

\begin{table}[htbp]
\centering
\small
\caption{Model families used in Sections~\ref{main_results} and~\ref{open-weighted-models} \citep{biderman2023pythia, black2021gptneo, peng2023rwkv, groeneveld2024olmo}.}
\label{tab:models}
\begin{tabular*}{\linewidth}{@{\extracolsep{\fill}}lccc@{}}
\toprule
\textbf{Family} & \textbf{Base Sizes} & \textbf{SFT} & \textbf{RL / Aligned} \\
\midrule
\multicolumn{4}{@{}l}{\textit{Corpus: The Pile}} \\
Pythia & 410M, 1.4B, 12B & 410M, 1.4B & 410M, 1.4B$^\dagger$ \\
GPT-Neo & 125M, 1.3B, 2.7B & -- & -- \\
RWKV-4 & 430M, 3B, 14B & 3B, 14B$^\ddagger$ & -- \\
\addlinespace
\multicolumn{4}{@{}l}{\textit{Corpus: Dolma}} \\
OLMo & 1B, 7B & 7B & 7B \\
\bottomrule
\multicolumn{4}{@{}l}{\scriptsize $^\dagger$Includes DPO \citep{rafailov2023dpo} (410M, 1.4B) and PPO \citep{schulman2017ppo} (1.4B). \enskip $^\ddagger$Raven checkpoints.}
\end{tabular*}
\end{table}

\subsection{Datasets}
\label{setup_datasets}
For each model family, we pair \emph{seen} subsets of its pre-training mixture with \emph{unseen} negative-control datasets.

\textbf{Seen Datasets (The Pile):} \texttt{github}, \texttt{wikipedia}, \texttt{stackexchange}, \texttt{pubmed\_abstracts}, \texttt{uspto\_backgrounds}, \texttt{cc}, \texttt{openwebtext2}, \texttt{pubmed\_central}, \texttt{freelaw}, \texttt{hackernews}, \texttt{nih\_exporter}, \texttt{arxiv}, \texttt{dm\_mathematics} \citep{gao2020pile}.

\textbf{Seen Datasets (Dolma):} \texttt{c4}, \texttt{wiki}, \texttt{pes2o}, \texttt{reddit}, \texttt{starcoder}, \texttt{cc} \citep{soldaini2024dolma}.

\textbf{Unseen Datasets (Shared):} \texttt{arxiv\_2025}, \texttt{bbc\_news\_2025}, \texttt{patent\_2025}, \texttt{gutenberg\_fiction}, \texttt{aime2025}, \texttt{aime2024}, \texttt{bbq} \citep{parrish2022bbq}, \texttt{reward\_bench\_v1} \citep{lambert2024rewardbench}, \texttt{frames} \citep{krishna2025frames}, \texttt{gpqa\_diamond} \citep{rein2024gpqa}, \texttt{amazon\_reviews\_2023} \citep{hou2024amazon}, \texttt{global\_news}, \texttt{livecodebench\_v1/v5} \citep{jain2025livecodebench}. (\textit{Note:} \texttt{gsm8k\_2021} \citep{cobbe2021gsm8k}, \texttt{math500\_2021} \citep{hendrycks2021math, lightman2024verify}, \texttt{humaneval\_2021} \citep{chen2021humaneval}, and \texttt{ifeval\_2023} \citep{zhou2023ifeval} were additionally used as unseen controls exclusively for Pile-trained models).

\subsection{Frontier Models and API Logistics}
\label{api_logistics}

We evaluate Claude Sonnet and Opus 4.5 \citep{anthropic2025sonnet45, anthropic2025opus45} via OpenRouter, DeepSeek-V3 and V4-Pro \citep{deepseekai2024v3, deepseekai2026v4} through identical API routing, and Kimi-k2.6 \citep{kimi2025k2, kimi2026k26} directly via Moonshot AI.

To isolate pure memorization from instruction-following bias, all requests use assistant prefill (M-mode; Appendix~\ref{modalify_shift}), forcing direct sequence continuation. For DeepSeek-V4-Pro, reasoning tokens are disabled. Rate limits and transient API failures are handled by an asynchronous execution wrapper with exponential backoff, preventing dropped queries.

Generation parameters are standardized with a 512-token cap and temperature 0 to minimize sampling stochasticity, except for Kimi-k2.6, which uses temperature 0.6 due to provider constraints. We target $n=100$ documents per dataset, with minor deviations: \texttt{pile\_dm\_mathematics} contains 99 valid documents, \texttt{aime2026} yields 25 after filtering items under 200 characters, and \texttt{reddit\_2026} provides 99 valid matches.

Using $K=3$ context documents, the theoretical ceiling is $W=16$ token positions ($\le 3{,}200$ calls per dataset), though realized volume scales down with length and overlap. Unconditioned completions regularly hit the 512-token cap, driving the completion volume. Table~\ref{tab:api_costs} summarizes token usage and cost across the evaluation suite.

\begin{table}[htbp]
\centering
\caption{Total API usage and costs aggregated across all dataset evaluations.}
\label{tab:api_costs}
\begin{tabular*}{\linewidth}{@{\extracolsep{\fill}}lrrr@{}}
\toprule
\textbf{Model} & \textbf{Prompt Tokens} & \textbf{Completion Tokens} & \textbf{Cost (USD)} \\
\midrule
Claude Opus 4.5 & 8,314,517 & 1,280,741 & \$73.59 \\
Claude Sonnet 4.5 & 8,314,517 & 853,024 & \$37.74 \\
DeepSeek-V4-Pro & 7,245,621 & 4,609,999 & \$9.74 \\
DeepSeek-V3 & 7,200,410 & 857,710 & \$3.07 \\
Kimi-k2.6 & 7,495,784 & 862,260 & $\le$ \$8.92 \\
\midrule
\textbf{Total} & & & $\le$ \textbf{\$133.06} \\
\bottomrule
\end{tabular*}
\end{table}
\newpage
\section{Additional Experiments}

\subsection{Context Construction Policy}
\label{context_policy}

To evaluate the robustness of our context construction approach, we ablate both the selection strategy ($\pi_c$) and the number of context documents ($K$). We measure performance using the separation gap $\Delta \mathcal{S} = \mathcal{S}_{\text{seen}} - \mathcal{S}_{\text{unseen}}$ and the detection AUC, averaged across two models (\texttt{gpt-neo-125m} and \texttt{pythia-410m}) under a strictly constrained token budget of $W=16$.

\begin{figure}[H]
    \centering
    \includegraphics[width=0.48\linewidth]{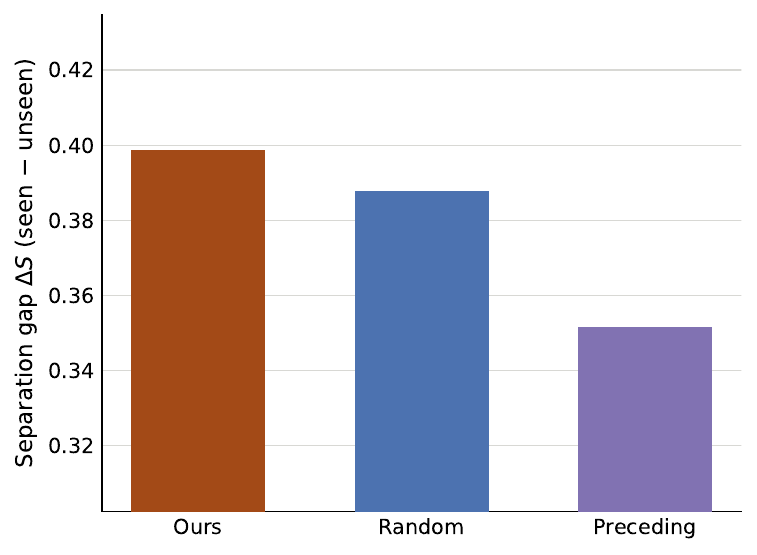}
    \hfill
    \includegraphics[width=0.48\linewidth]{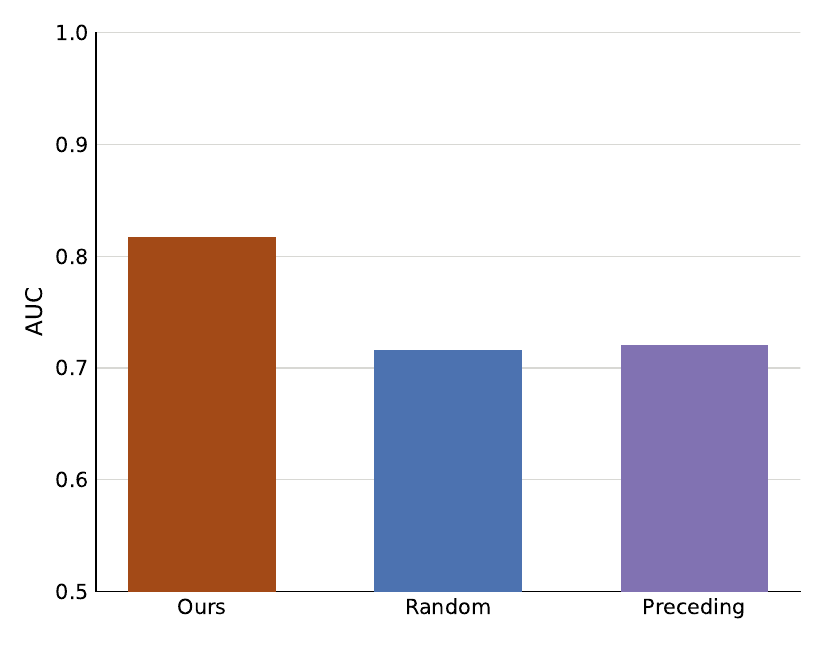}
    \caption{Context selection policy ablation. \textbf{Left:} Mean separation gap ($\Delta \mathcal{S}$). \textbf{Right:} Contamination-detection AUC. Both metrics use $K=3$ context documents, $W=16$ selected tokens, and are averaged over \texttt{gpt-neo-125m} and \texttt{pythia-410m}.}
    \label{fig:context_policy_ablation}
\end{figure}

\paragraph{Context Selection Strategy.}
We compare three methods for constructing the prepended context $C$: our default \textit{similar-prefixes} policy (Jaccard-ranked) \citep{jaccard1912distribution}, \textit{random prefixes} \citep[cf.][]{codec}, and \textit{preceding} (sequential text). As shown in Figure~\ref{fig:context_policy_ablation}, our Jaccard-based selection provides a large advantage over both naive baselines. It yields the highest mean separation gap ($\Delta \mathcal{S} \approx 0.399$) and substantially boosts the detection performance to an AUC of $0.817$. In contrast, random and sequential-preceding selections perform nearly identically in terms of AUC ($0.716$ and $0.721$, respectively), a drop of roughly $0.10$ AUC ($\approx 14\%$ relative) compared to our method. This confirms that dynamically maximizing lexical overlap is crucial for recovering a strong contamination signal \citep[cf.][]{liu2022makes}.

\begin{wrapfigure}{r}{0.45\textwidth}
    \centering
    \vspace{-15pt}
    \includegraphics[width=\linewidth]{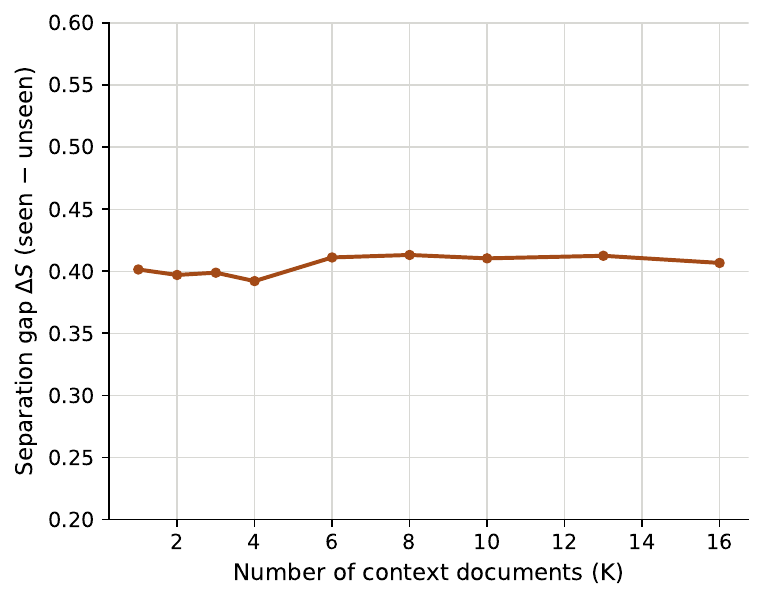}
    \vspace{-10pt}
    \caption{The effect of varying the number of context documents ($K$) on the separation gap.}
    \label{fig:context_k_gap}
\end{wrapfigure}

\paragraph{Context Size ($K$).}
Having established that the choice of documents is critical, we evaluate the sensitivity of the signal to the volume of context. We fix the selection policy to \textit{similar-prefixes} and vary the number of supplied context documents $K$ from 1 to 16. Figure~\ref{fig:context_k_gap} demonstrates that the separation gap remains highly stable, fluctuating within a narrow band of $0.392$ to $0.413$. Providing a single context document ($K=1$, $\Delta \mathcal{S} = 0.401$) recovers essentially the exact same membership signal as providing sixteen documents ($K=16$, $\Delta \mathcal{S} = 0.407$). This indicates that the separation power is highly sensitive to \textit{which} documents are selected, but largely insensitive to \textit{how many} are supplied. Thus, $K$ acts mainly as a compute and cost parameter rather than an accuracy bottleneck.

\subsection{Non-Zero Temperature Sampling}

To evaluate the robustness of our framework when access to deterministic decoding is restricted, we sweep the sampling temperature from 0 to 1. We hold the audit configuration fixed at our optimal parameters ($K=3$, $W=16$, LCS policy) and average the results across six models (Pythia 12B, 1.4B, 410M, and GPT-Neo 2.7B, 1.3B, 125M) \citep{biderman2023pythia, black2021gptneo}.

\begin{figure}[H]
    \centering
    \includegraphics[width=0.48\linewidth]{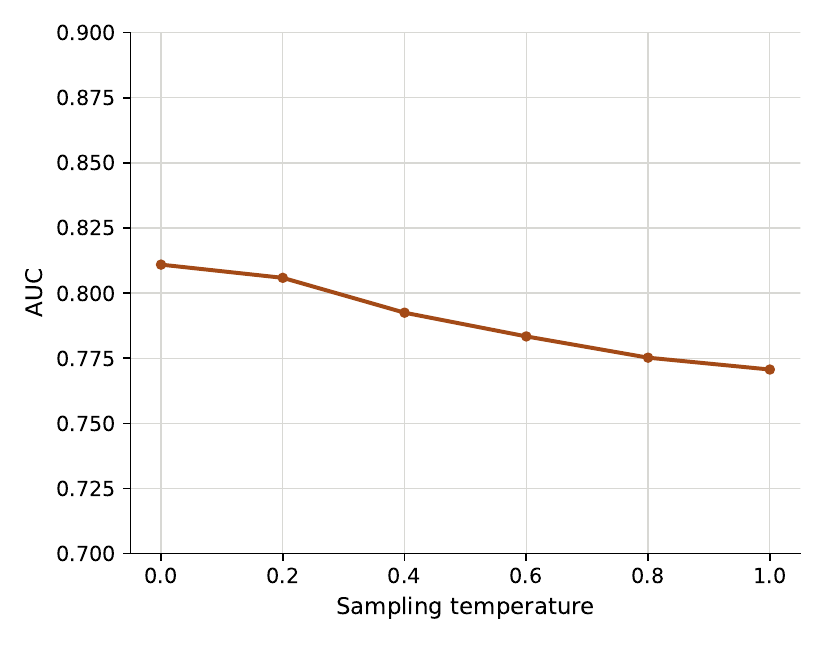}
    \hfill
    \includegraphics[width=0.48\linewidth]{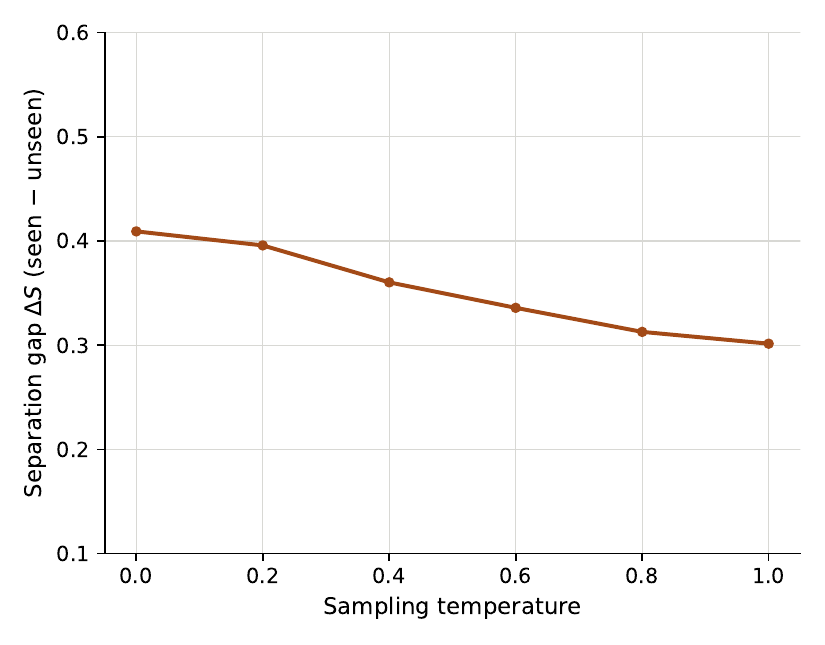}
    \caption{Detection robustness under non-zero temperature sampling. \textbf{Left:} Contamination-detection AUC as temperature varies. \textbf{Right:} Mean score separation gap ($\Delta \mathcal{S}$) between contaminated and held-out corpora. Results are averaged over 6 models.}
    \label{fig:temperature_sweep}
\end{figure}

As shown in Figure~\ref{fig:temperature_sweep}, contamination detection degrades gracefully, rather than catastrophically, as the sampling temperature increases. The detection AUC drops by only $5\%$ relative, from $0.811$ at temperature 0 to $0.771$ at temperature 1. Concurrently, the raw separation gap ($\Delta \mathcal{S}$) shrinks more noticeably, decreasing by roughly $26\%$ relative (from $0.409$ to $0.301$) over the same temperature range. These results demonstrate that while stochastic sampling compresses the margin between seen and unseen distributions, the method maintains strong discriminative power (AUC $\gg 0.5$) even under fairly aggressive temperature settings.

\subsection{Dataset-Level False Positives}
In table \ref{tab:lomo_fp}, we fix a threshold for each model using only the other models' unseen datasets (at most 5\% flagged). LCS wrongly flags 9 of 179 unseen datasets (6 of them \texttt{patent\_2025}) and detects 116 of 129 seen ones (89.9\%). Most misses are \texttt{pile\_arxiv}. At the same false-alarm level, the black-box baselines detect at most 51 of 129. On the 8 post-trained models, LCS flags 8 of 128 unseen datasets and detects 81 of 90 seen ones.

\begin{table}[b]
\centering
\caption{Dataset-level false positives at a leave-one-model-out threshold on the 11 base models. For each model, the threshold is set to give at most 5\% FPR on the other models' unseen datasets and applied to the held-out model. $^*$\texttt{global\_news} is excluded for these methods, which require at least 100 tokens per document.}
\label{tab:lomo_fp}
\small
\begin{tabular*}{\linewidth}{@{\extracolsep{\fill}}llcc@{}}
\toprule
Method & Access & Unseen flagged (FP) & Seen detected (TP) \\
\midrule
LCS, $W{=}16$ (ours) & black-box & 9 / 179 (5.0\%) & \textbf{116 / 129 (89.9\%)} \\
Carlini$^*$ & black-box & 8 / 168 (4.8\%) & 51 / 129 (39.5\%) \\
SA-MIA$^*$ & black-box & 10 / 168 (6.0\%) & 35 / 129 (27.1\%) \\
SimMIA & black-box & 11 / 179 (6.1\%) & 16 / 129 (12.4\%) \\
\midrule
Loss & gray-box & 11 / 179 (6.1\%) & 21 / 129 (16.3\%) \\
Min-K\% & gray-box & 10 / 179 (5.6\%) & 16 / 129 (12.4\%) \\
zlib & gray-box & 9 / 179 (5.0\%) & 26 / 129 (20.2\%) \\
CoDeC & gray-box & 9 / 179 (5.0\%) & 110 / 129 (85.3\%) \\
\bottomrule
\end{tabular*}
\end{table}

\subsection{Per-Document Score Comparison on Public Models}
In Tables \ref{tab:table_ots_auc_doc}, \ref{tab:post_train_doc} we present AUC and TPR@5\%FPR scores when pooled over documents, complementing to what was presented in Tables \ref{tab:table_ots_auc_blackbox_full}, \ref{tab:table_ots_auc}, \ref{tab:post_train} (where scores were pooled over the entire datasets). 

At the document level, where each document is scored on its own with at most 32
calls, LCS separates seen from unseen less sharply than at the dataset level (mean AUC
0.780 vs.\ 0.952), and its position among the gray-box methods changes
(Table~\ref{tab:table_ots_auc_doc}): it stays above Loss and Min-K\% and matches zlib
in AUC (0.780 vs.\ 0.771), but zlib has a higher TPR@5\%FPR (0.279 vs.\ 0.173) and
CoDeC is ahead on both metrics (0.856 and 0.350). LCS still exceeds the black-box
baselines (mean AUC 0.542--0.558; Table~\ref{tab:ext-auc}). Across post-training
(Table~\ref{tab:post_train_doc}), AUC drops by at most 0.03 except for RWKV-4 3B after
SFT (0.792 to 0.722). TPR@5\%FPR is low and non-monotonic across checkpoints (e.g.,
Pythia 1.4B: 0.158, 0.240, 0.132 for base, SFT, DPO), so we do not interpret its
changes.

\begin{table}[t]
\centering
\small
\begin{minipage}[t]{0.31\linewidth}
\centering
\caption{LCS vs.\ gray-box methods, per document.}
\label{tab:table_ots_auc_doc}
\setlength{\tabcolsep}{4pt}
\begin{tabular}{@{}lcc@{}}
\toprule
& \multicolumn{2}{c}{Per document} \\
\cmidrule(lr){2-3}
Method & AUC & TPR \\
\midrule
Loss & 0.589 & 0.129 \\
Min-K\% & 0.580 & 0.118 \\
zlib & 0.771 & 0.279 \\
CoDeC & \textbf{0.856} & \textbf{0.350} \\
LCS (Ours) & 0.780 & 0.173 \\
\bottomrule
\end{tabular}
\end{minipage}%
\hfill
\begin{minipage}[t]{0.65\linewidth}
\centering
\caption{Post-trained checkpoints, pooled over documents. Aligned: DPO for Pythia, Instruct for OLMo. TPR at 5\% FPR.}
\label{tab:post_train_doc}
\setlength{\tabcolsep}{3.5pt}
\begin{tabular}{@{}lcccccc@{}}
\toprule
& \multicolumn{2}{c}{Base} & \multicolumn{2}{c}{+ SFT} & \multicolumn{2}{c}{+ Aligned} \\
\cmidrule(lr){2-3}\cmidrule(lr){4-5}\cmidrule(lr){6-7}
Model & AUC & TPR & AUC & TPR & AUC & TPR \\
\midrule
Pythia 410M & 0.811 & 0.289 & 0.805 & 0.143 & 0.801 & 0.148 \\
Pythia 1.4B & 0.816 & 0.158 & 0.803 & 0.240 & 0.797 & 0.132 \\
RWKV-4 3B   & 0.792 & 0.214 & 0.722 & 0.079 & --    & --    \\
RWKV-4 14B  & 0.724 & 0.051 & 0.725 & 0.051 & --    & --    \\
OLMo 7B     & 0.717 & 0.085 & 0.688 & 0.110 & 0.700 & 0.117 \\
\bottomrule
\end{tabular}
\end{minipage}
\end{table}

\subsection{Randomized Exposure}\label{app:finetuning_noise}

Temporal splits cannot by themselves separate membership from dataset shift. To test whether the score tracks training exposure when membership is independent of the data distribution, we construct a controlled setup in which models are fine-tuned for a single epoch on randomly chosen subsets of a common pool of corpora. Building upon the architectures evaluated in Appendix~\ref{setup_models}, we selected an expanded set of 14 Pile-pretrained models: Pythia (160M, 410M, 1.4B, 2.8B, 6.9B, 12B) \citep{biderman2023pythia}, GPT-Neo (125M, 1.3B, 2.7B) \citep{black2021gptneo}, GPT-J (6B), and RWKV-4 (430M, 3B, 7B, 14B) \citep{peng2023rwkv}. Additionally, we constructed a contamination pool of 16 corpora released after the Pile knowledge cutoff \citep{gao2020pile}. While largely overlapping with the negative controls established in Appendix~\ref{setup_datasets}, this pool was specifically tailored for this experiment: arXiv abstracts, BBC news, USPTO patents, Project Gutenberg fiction, GSM8K, MATH-500, HumanEval, IFEval, BBQ, RewardBench, FRAMES, Amazon reviews, global news, LiveCodeBench, HH-RLHF-helpful \citep{bai2022hh}, and TinyStories \citep{eldan2023tinystories}. Each corpus was capped at 600 documents. We drew 4 random seen/unseen partitions of this pool (7--10 seen corpora each) to contaminate the models via single-epoch fine-tuning. Models $\ge 2$B parameters used merged LoRA adapters ($r=16$) \citep{hu2022lora}, while smaller models were fully fine-tuned, yielding 56 unique checkpoints. 

Membership is assigned at random, independently of domain, format and date. The partitions are not balanced, so some corpora are seen more often than others. This is a constructed setting that trades realism for control over membership. The models are
fine-tuned directly on a small set of corpora, so exposure is concentrated and we expect the seen data to be fitted more strongly than in pre-trained models, where a corpus is only a small part of a much larger training mix. This should make seen data easier to detect for methods that read out fitting, such as likelihood-based scores. The main experiments
(Section~\ref{open-weighted-models}) use pre-trained models and their actual training data. 

We audited each checkpoint on all 16 corpora ($n=100$ documents per corpus, $K=3$ context documents) under 5 window-selection policies and 9 window budgets $W \in \{5, \dots, 300\}$, totaling 2,520 audits.

\begin{wrapfigure}{r}{0.45\textwidth}
    \vspace{-1.5em}
    \centering
    \includegraphics[width=\linewidth]{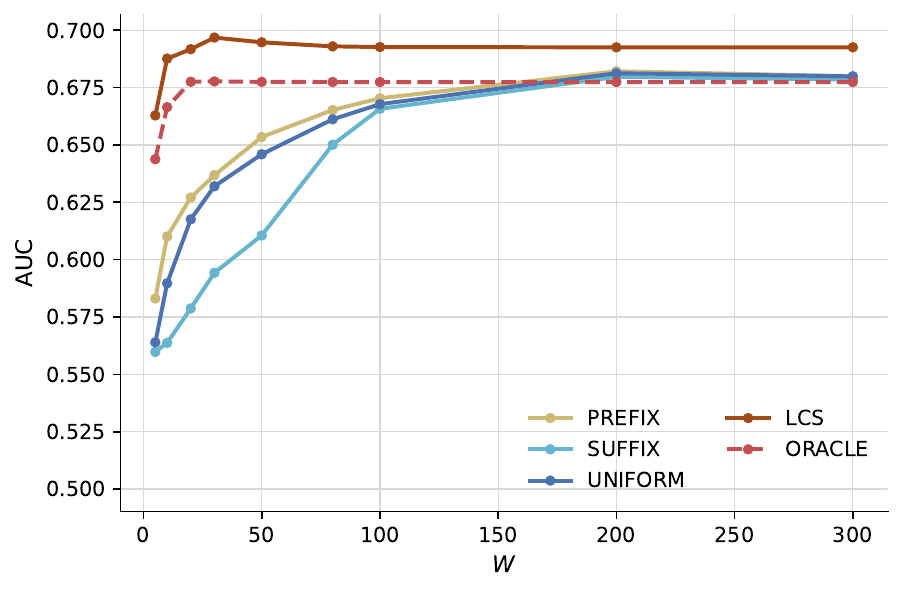}
    \caption{AUC performance of selection policies in random-exposure setting. The LCS policy shifts flip balance and outperforms the indiscriminate Oracle aggregation.}
    \label{fig:finetuning_noise_tradeoff}
    \vspace{-1em}
\end{wrapfigure}

In this setting, LCS reaches a higher AUC than the Oracle policy, which aggregates all
positions with $\delta_j \neq 0$, at all budgets $W$
(Figure~\ref{fig:finetuning_noise_tradeoff}). The Oracle selects positions using the
outcome the score is computed from, so it is a reference for the density of informative
positions and not an upper bound on AUC. Appendix~\ref{app:density_sign} compares what the policies select: LCS picks positions that flip more often than Uniform and shifts the balance of helpful
vs.\ harmful flips.

\begin{table}[t]
\centering
\caption{Comparison with baselines in the randomized-exposure setting. Metrics are computed per checkpoint and averaged over the 56 checkpoints. Dataset-level scores are pooled over documents within each dataset. TPR is at 5\% FPR.}
\label{tab:c3-baselines}
\begin{tabular*}{0.9\linewidth}{@{\extracolsep{\fill}}llcccc@{}}
\toprule
& & \multicolumn{2}{c}{Document level} & \multicolumn{2}{c}{Dataset level} \\
\cmidrule(lr){3-4} \cmidrule(lr){5-6}
Method & Access & AUC & TPR & AUC & TPR \\
\midrule
LCS, $W{=}16$ (ours) & black-box & 0.706 & 0.120 & 0.858 & 0.680 \\
LCS, $W{=}32$ (ours) & black-box & 0.714 & 0.144 & 0.844 & 0.625 \\
\midrule
Carlini & black-box & 0.571 & 0.090 & 0.757 & 0.310 \\
SA-MIA  & black-box & 0.738 & 0.218 & 0.835 & 0.542 \\
SimMIA  & black-box & 0.531 & 0.067 & 0.607 & 0.183 \\
\midrule
Loss    & gray-box  & 0.808 & 0.390 & 0.849 & 0.554 \\
Min-K\% & gray-box  & \textbf{0.835} & \textbf{0.477} & 0.879 & 0.622 \\
zlib    & gray-box  & 0.555 & 0.053 & 0.550 & 0.008 \\
CoDeC   & gray-box  & 0.770 & 0.170 & \textbf{0.951} & \textbf{0.796} \\
\bottomrule
\end{tabular*}
\end{table}

Table~\ref{tab:c3-baselines} compares LCS with the main baselines
in this setting. At the dataset level, LCS ($W{=}16$) reaches AUC 0.858: above the other
black-box baselines (Carlini 0.757, SA-MIA 0.835), on par with the gray-box Loss (0.849)
and above zlib (0.550), but below Min-K\% (0.879) and CoDeC (0.951); its TPR@5\%FPR
(0.680) is second only to CoDeC (0.796). At the document level, SA-MIA (0.738) and the
gray-box Loss, Min-K\% and CoDeC (0.770--0.835) are ahead of LCS (0.706). Doubling the
budget to $W{=}32$ does not change this picture (dataset-level AUC 0.844).

Compared with the pre-trained models of Section~\ref{open-weighted-models} (different
models and corpora, so only qualitatively comparable), the ranking changes. Loss and
Min-K\% improve strongly (0.585 and 0.599 there), as expected when the model has just
been fitted to the seen data, whereas zlib falls to chance (0.759 there), which suggests
that part of its advantage under temporal splits reflects differences between the
corpora rather than membership. LCS (0.952 there) and CoDeC (0.961 there) remain well
above chance.

\subsection{Increasing Exposure}

\begin{wrapfigure}{r}{0.42\textwidth}
    \centering
    \vspace{-12pt}
    \includegraphics[width=\linewidth]{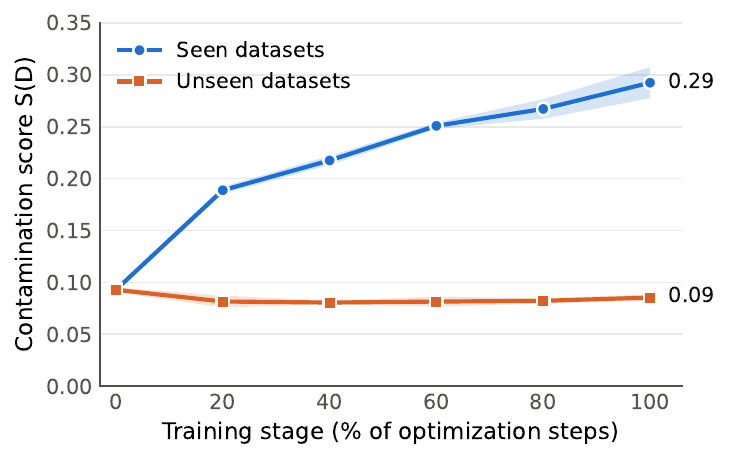}
    \vspace{-16pt}
    \caption{Scores $\mathcal S(\mathcal D_{\text{seen}})$ and $\mathcal S(\mathcal D_{\text{unseen}})$ across training stages.}
    \label{fig:calib}
    \vspace{-10pt}
\end{wrapfigure}

We fine-tune Pythia-410M on each of the four seen/unseen partitions of
Appendix~\ref{app:finetuning_noise} and audit intermediate checkpoints
throughout the single epoch. Figure~\ref{fig:calib} shows that, as exposure to
the seen corpora grows, the mean score on seen datasets 
$\mathcal{S}(\mathcal{D}_{\text{seen}})$ rises while the mean score on unseen datasets stays flat. For Pythia-410M the score therefore tracks the amount of exposure, not only the seen/unseen label. Before any fine-tuning the two groups score
identically (0.093 vs.\ 0.093), hence the gap that opens during training (0.000 to
0.207) cannot come from differences between the two groups of datasets.

\subsection{Blind Baselines}
\label{apd:blind_baselines}

Following \citet{das2025blind}, we measure how well a model-free classifier recovers the labels from text alone, and test directly whether the LCS score tracks training exposure rather than such differences.

\begin{wraptable}{r}{0.4\textwidth}
\vspace{-1.2em}
\centering\small
\caption{Blind baseline vs.\ LCS on the main experiments (dataset level).}
\label{tab:blind}
\begin{tabular*}{\linewidth}{@{\extracolsep{\fill}}lcc@{}}
\toprule
Method & AUC & TPR@5\%FPR \\
\midrule
Blind classifier & 0.718 & 0.000 \\
LCS, $W{=}16$    & 0.952 & 0.893 \\
\bottomrule
\end{tabular*}
\vspace{-1em}
\end{wraptable}

\paragraph{Main Results.}

We set as a classifier a logistic regression on TF-IDF n-gram features of the documents (word 1–2-grams and character 3–5-grams), trained using leave-one-dataset-out to classify between seen and unseen data from the main setup (Section \ref{open-weighted-models}). It uses no model outputs. We then compute dataset-level AUC and TPR@5\%FPR over datasets, exactly as for LCS.

The blind classifier recovers the labels better than chance (Table~\ref{tab:blind}), so the datasets are partly separable by text, as expected for temporal splits. At the 5\% FPR operating point, however, it detects none of the seen datasets, whereas LCS is far higher (0.952 AUC, 0.893 TPR). The blind AUC measures how separable the labels are, which is a property of the benchmark; it does not show whether a score exploits that separability; the next experiment tests whether LCS uses data shifts to its advantage.

\paragraph{Dataset Separability Usage.}

We take $D_1$ = GSM8K and $D_2$ = MBPP, which a text classifier separates perfectly, and fine-tune Pythia-410M for 4 epochs in four scenarios: (A) neither dataset, (B) both, (C1) only $D_1$, (C2) only $D_2$.

As Table \ref{tab:twodata} shows, with equal exposure (A, B) LCS does not separate the datasets, although the text classifier does so perfectly. In C1 and C2 the datasets are separated and the ranking of $s(D_1)$ and $s(D_2)$ flips, while the text-based AUC stays the same. The score therefore follows exposure and not text differences.

\begin{table}[h]
\centering
\caption{Two-dataset exposure experiment. 
$\dagger$: the AUC of the text-only classifier for separating $D_1$ from $D_2$. $\ddagger$: oriented toward the more exposed dataset.}
\label{tab:twodata}
\begin{tabular*}{\linewidth}{@{\extracolsep{\fill}}lcccc@{}}
\toprule
Scenario & Classifier AUC$^\dagger$ & $s(D_1)$ & $s(D_2)$ & LCS AUC$^\ddagger$ \\
\midrule
A (neither)     & 1.000 & 0.071 & 0.056 & 0.511 \\
B (both)        & 1.000 & $0.626 \pm 0.047$ & $0.602 \pm 0.055$ & $0.529 \pm 0.039$ \\
C1 (only $D_1$) & 1.000 & $0.533 \pm 0.066$ & $0.067 \pm 0.004$ & $0.949 \pm 0.014$ \\
C2 (only $D_2$) & 1.000 & $0.074 \pm 0.010$ & $0.638 \pm 0.034$ & $0.960 \pm 0.008$ \\
\bottomrule
\end{tabular*}
\end{table}

\subsection{M-Mode vs.\ I-Mode Prompting}\label{modalify_shift}

We evaluate audits that wrap each query in a chat template \citep{ouyang2022training}, which we call I-mode:
\begin{equation}
    i_{C}^j = \mathbf{1}\bigl[\hat x_j = x_j \mid \texttt{<user>}\, C, x_{<j}\, \texttt{</user>}\, \texttt{<assistant>}\bigr],
\end{equation}
yielding the indicator $\delta_j = i_{C}^j - i^j_\emptyset$. We also test assistant prefill \citep{qi2025safety}, which we call M-mode:
\begin{equation}
    i_{C}^j = \mathbf{1}\bigl[\hat x_j = x_j \mid \texttt{<assistant>}\, C, x_{<j}\bigr],
\end{equation}
and compare both with prompting without any template (U-mode), as used in Sections~\ref{main_results} and~\ref{open-weighted-models}.

\begin{table}[t]
\centering
\caption{Fraction of target tokens ranked within the top-$K$ predictions under U-mode, M-mode, and I-mode. The $K{=}1$ column corresponds to the audit's rule.}
\label{tab:target_token_topk}
\small
\setlength{\tabcolsep}{4pt}
\begin{tabular*}{\linewidth}{@{\extracolsep{\fill}}lccccccccccc@{}}
\toprule
Top-$K$ & 1 & 2 & 3 & 5 & 10 & 20 & 50 & 100 & 500 & 1000 & 10000 \\
\midrule
U-mode & 0.741 & 0.805 & 0.836 & 0.859 & 0.886 & 0.914 & 0.950 & 0.965 & 0.988 & 0.995 & 0.999 \\
M-mode & 0.726 & 0.796 & 0.833 & 0.857 & 0.886 & 0.911 & 0.948 & 0.964 & 0.990 & 0.995 & 0.999 \\
I-mode & \textbf{0.016} & 0.018 & 0.022 & 0.028 & 0.086 & 0.133 & 0.197 & 0.230 & 0.298 & 0.349 & 0.715 \\
\bottomrule
\end{tabular*}
\end{table}

For five randomly sampled datasets, Table~\ref{tab:target_token_topk} shows that U-mode and M-mode place the target token at top-1 in about 73\% of cases, whereas I-mode does so in under 2\% and shifts the entire distribution.

This failure stems likely from instruction framing inducing distinct operational modes \citep{lin2024unlocking}. With prompt template $\chi$, a language model can be viewed as the marginalization
\begin{equation}
    p(x_{t} \mid x_{<t}, \chi) = \sum_z p_z(x_t \mid x_{<t})\, p(z \mid \chi)
\end{equation}
over modes $z$ \citep[cf.][]{xie2022explanation}. We hypothesize that in I-mode, $p_{\text{I}}$ dominates, so $D_{KL}(p_{\text{M}} \parallel p_{\text{I}})$ is high, whereas $D_{KL}(p_{\text{M}} \parallel p_{\text{U}})$ remains low as both model standard language. Where prefill is supported, M-mode would thus faithfully proxy raw continuation, consistent with alignment leaving base-model behavior largely intact \citep{nasr2025scalable, lin2024unlocking}.

\section{Details of the LCS Heuristic}\label{lcs_details}
Given a document $x=(x_1, \dots, x_T)$ and the context $C$, LCS selects the positions at which $C$ gives the strongest literal evidence for the next token. For each position $j$, define the match length
\[
\ell_j = \max\bigl\{ r \ge 0 : (x_{j-r+1},\dots, x_j) \text{ occurs as a contiguous subsequence in } C \bigr\}.
\]
A position $j$ is a candidate if $\ell_j \ge \ell_{\min} = 2$ and $j > s = 10$, where $s$ is the number of initial tokens skipped. Let $m$ be the number of candidates and $W$ the window budget.
\begin{itemize}
    \item $\boldsymbol{m > W}$: candidates are ranked by $\ell_j$ in descending order, with ties broken by earlier position in the document (i.e., smaller $j$). For instance, a 5-token match is chosen before a trigram match, and a trigram before a bigram. The top $W$ positions are then queried.
    \item $\boldsymbol{m \le W}$: all $m$ candidates are queried in document order.
    \item $\boldsymbol{m = 0}$: the document has no window and is excluded from scoring.
\end{itemize}

\subsection{Density and Sign}\label{app:density_sign}
Table~\ref{tab:flip-polarity} compares the positions chosen by LCS, Uniform and the Oracle in the randomized-exposure setting of Appendix~\ref{app:finetuning_noise}. LCS differs from Uniform in two ways. First, \textbf{density}: a larger fraction of the positions selected by LCS flip ($\delta \neq 0$), $14.2\%$ on seen and $25.3\%$ on unseen data versus $11.6\%$ and $14.6\%$ for Uniform ($1.2\times$ and $1.7\times$), but far below the Oracle's $100\%$. Second, \textbf{sign balance}: relative to Uniform, LCS over-represents positions where the context helps. Among the flipped positions, the context-hurt share, which is the quantity underlying the score $\mathcal{S}$, is $20.4\%$ on seen and $7.4\%$ on unseen data for LCS, versus $35.7\%$ and $20.5\%$ for Uniform. Under both policies this share is higher on seen than on unseen data, with a similar difference ($13.0$ vs.\ $15.2$ points), so LCS retains the seen--unseen contrast while collecting more flips per query. Since LCS shifts the sign balance, it changes the quantity the score estimates relative to Uniform.

\begin{table}[h]
\centering
\caption{Which positions each selection policy picks, on seen and unseen data. Flip rate is the share of selected positions with $\delta \neq 0$ (the Oracle selects flipped positions by construction). \emph{Context helped} counts wrong$\to$right and \emph{context hurt} right$\to$wrong flips. All values in \%.}
\label{tab:flip-polarity}
\begin{tabular*}{\linewidth}{@{\extracolsep{\fill}}lcccccc@{}}
\toprule
& \multicolumn{2}{c}{Flip rate} & \multicolumn{2}{c}{Context helped} & \multicolumn{2}{c}{Context hurt} \\
& \multicolumn{2}{c}{(of selected)} & \multicolumn{2}{c}{(of selected)} & \multicolumn{2}{c}{(of flipped)} \\
\cmidrule(lr){2-3} \cmidrule(lr){4-5} \cmidrule(lr){6-7}
Policy & Seen & Unseen & Seen & Unseen & Seen & Unseen \\
\midrule
Oracle  & 100.0 & 100.0 & 55.1 & 69.6 & 44.9 & 30.4 \\
LCS     & 14.2  & 25.3  & 11.3 & 23.5 & 20.4 & 7.4  \\
Uniform & 11.6  & 14.6  & 7.5  & 11.6 & 35.7 & 20.5 \\
\bottomrule
\end{tabular*}
\end{table}

\subsection{Accuracy and Token Selection}

In this section we compare the LCS policy at $W{=}16$ to two accuracy baselines
that use no context. Let $a_j=1[\hat x_j = x_j \mid x_{<j}]$ be the
unconditioned top-1 indicator at position $j$, and let $\mathcal{P}_\pi(x)$ be
the set of $W$ positions selected by a policy $\pi$. Both baselines score a
document by its accuracy on the selected positions,
\[
  Acc_\pi(x) \;=\; \frac{1}{|\mathcal{P}_\pi(x)|}\sum_{j\in\mathcal{P}_\pi(x)} a_j ,
\]
and dataset scores average $Acc_\pi$ over its documents, exactly as for
our score (higher value corresponds to higher contamination as in the case of our method).

\begin{itemize}
  \item $Acc_{\text{LCS}}$ uses the LCS policy for position selection. This baseline isolates the effect of the context comparison from the effect of token selection.
  \item $Acc_{\text{prefix}}$ computes the accuracy on the first $W$ positions of the document, which isolates the effect of token selection itself.
\end{itemize}
Importantly, our score $\Delta_{bb}$ compares each position with and without context, while
these baselines use only the unconditioned prediction accuracy.

To make the comparison fair budget-wise, our score uses two queries per position, i.e., $W=16$ costs 32 calls. An accuracy baseline at $W=16$ costs 16 calls (one call per position). We therefore include in the comparison $Acc_\pi$ with $W=32$.

As observed in Table \ref{tab:acc_baselines} accuracy at LCS positions is a strong baseline, and token selection matters:
it exceeds accuracy at prefix positions for base/post-trained models (e.g., dataset-level AUC
$0.771$ vs.\ $0.687$ at $W{=}16$ on base models). Doubling the $Acc_{\pi}$ budget
does not close the gap to our score. At equal calls, comparing $Acc_\text{LCS}$
at $W{=}32$ with our score, AUC rises from $0.682$ to $0.780$ at the document
level and from $0.791$ to $0.952$ at the dataset level on base models ($0.709$
to $0.942$ on post-trained models).

In the randomized-exposure setting of Appendix~\ref{app:finetuning_noise} the ranking reverses: accuracy exceeds our score (dataset-level AUC $0.923$--$0.935$ vs.\ $0.858$ at 32 calls), and selecting LCS positions brings no gain over prefix positions at the dataset level (AUC $0.897$ vs.\ $0.918$ at $W{=}16$). This is expected as these models are trained directly on the seen datasets, which by design raises their no-context accuracy on exactly these datasets, so accuracy measures how well the model has fitted them. On pre-trained models a corpus is only a small part of a much larger training mix, and accuracy is much weaker. We therefore do not claim that the context contrast outperforms accuracy in every regime: it helps where no-context accuracy is a weak signal, as for the pre-trained models we audit, while Appendix~\ref{app:finetuning_noise} serves to show that the score tracks exposure when membership is randomized.

\begin{table}[bh]
\centering
\caption{Accuracy baselines vs.\ our score (LCS, $W{=}16$). Document-level
metrics pool documents; dataset-level metrics are computed per model and
averaged over models. Calls/doc is the number of queries per document.}
\label{tab:acc_baselines}
\small
\begin{tabular*}{\linewidth}{@{\extracolsep{\fill}}llccccc@{}}
\toprule
 & & Calls & \multicolumn{2}{c}{Document level} & \multicolumn{2}{c}{Dataset level} \\
\cmidrule(lr){4-5}\cmidrule(lr){6-7}
 & Method & /doc & AUC & TPR@5\% & AUC & TPR@5\% \\
\midrule
Base & $Acc_{\text{prefix}}$ $W{=}16$ & 16 & 0.610 & 0.141 & 0.687 & 0.273 \\
                & $Acc_{\text{prefix}}$ $W{=}32$ & 32 & 0.628 & 0.182 & 0.674 & 0.344 \\
                & $Acc_{\text{LCS}}$ $W{=}16$    & 16 & 0.682 & 0.141 & 0.771 & 0.393 \\
                & $Acc_{\text{LCS}}$ $W{=}32$    & 32 & 0.682 & \textbf{0.215} & 0.791 & 0.441 \\
                & LCS & 32 & \textbf{0.780} & 0.173 & \textbf{0.952} & \textbf{0.893} \\
\midrule
Post-trained & $Acc_{\text{prefix}}$ $W{=}16$ & 16 & 0.577 & 0.121 & 0.617 & 0.250 \\
                & $Acc_{\text{prefix}}$ $W{=}32$ & 32 & 0.592 & 0.145 & 0.603 & 0.240 \\
                & $Acc_{\text{LCS}}$ $W{=}16$    & 16 & 0.647 & 0.127 & 0.699 & 0.308 \\
                & $Acc_{\text{LCS}}$ $W{=}32$    & 32 & 0.644 & \textbf{0.193} & 0.709 & 0.327 \\
                & LCS & 32 & \textbf{0.755} & 0.127 & \textbf{0.942} & \textbf{0.901} \\
\midrule
Randomized & $Acc_{\text{prefix}}$ $W{=}16$ & 16 & 0.746 & 0.259 & 0.918 & 0.719 \\
 & $Acc_{\text{prefix}}$ $W{=}32$ & 32 & \textbf{0.787} & \textbf{0.309} & \textbf{0.935} & 0.729 \\
 & $Acc_{\text{LCS}}$ $W{=}16$    & 16 & 0.749 & 0.160 & 0.897 & 0.595 \\
 & $Acc_{\text{LCS}}$ $W{=}32$    & 32 & 0.781 & 0.243 & 0.923 & \textbf{0.755} \\
 & LCS & 32 & 0.706 & 0.120 & 0.858 & 0.680 \\
\bottomrule
\end{tabular*}
\end{table}

\newpage
\section{Extended Audit Results}\label{heatmaps}

\subsection{Public Models}
This section extends Section~\ref{open-weighted-models} with per-model AUC and TPR@5\%FPR, pooled over documents (Tables~\ref{tab:ext-auc} and~\ref{tab:ext-tpr}) and over datasets (Tables~\ref{tab:ext-auc-ds} and~\ref{tab:ext-tpr-ds}), all with 95\% CIs, for Carlini \citep{carlini2023quantifying}, SA-MIA \citep{kaneko2024samplingbasedpseudolikelihoodmembershipinference}, SimMIA \citep{yi2026simmia}, and LCS. LCS outperforms all baselines in AUC on every model, at both levels. At the document level, SA-MIA occasionally attains a higher TPR@5\%FPR, likely by capturing low-entropy documents, but its near-chance AUC limits its overall reliability; at the dataset level, LCS has the highest TPR@5\%FPR on every model.

\subsection{Reliability of the Claude Audits}\label{reliability}
All Claude queries use assistant prefill. Without context, 92.9\% (Opus 4.5) and 93.8\% (Sonnet 4.5) of replies are written in the assistant’s voice (I-mode) instead of continuing the document, and the target token is reproduced at only 0.50\% and 1.24\% of positions. Of all flips, 90\% (Opus) and 95\% (Sonnet) have an assistant-voice reply on the incorrect side. The scores therefore mostly measure which condition leaves continuation mode, not memorization, and we do not interpret them as contamination evidence. Additional failures we observe:
(1) Context triggers commentary: all 19 $n_-$ flips behind Opus’s $\mathcal S=0.95$ on \texttt{pile\_pubmed\_central} have a reformatted or assistant-voice reply with context. Sonnet’s $\mathcal S=0.11$ on \texttt{pile\_wikipedia} comes from no-context replies breaking. (2) Opus’s $\mathcal S=1.00$ on MATH-500 comes from three flips in one document. (3) Empty replies (112 on GPQA Diamond for Sonnet; none for Opus). (4) AIME yields no flips for either of the models.
For better interpretability, in Table \ref{tab:api_scores} we present contamination scores with confidence intervals.

\begin{table}[h!]
\scriptsize
\centering
\caption{Contamination scores (95\% intervals) per dataset and model. n/a: score undefined because no flips were observed.}
\label{tab:api_scores}
\small
\setlength{\tabcolsep}{3pt}
\resizebox{\linewidth}{!}{%
\begin{tabular}{@{}lccccc@{}}
\toprule
Dataset & Kimi K2.6 & DeepSeek V3 & DeepSeek V4 Pro & Claude Sonnet 4.5 & Claude Opus 4.5 \\
\midrule
\texttt{mmlu} & 0.21 {\scriptsize[0.16,0.28]} & 0.25 {\scriptsize[0.16,0.34]} & 0.34 {\scriptsize[0.27,0.42]} & 0.00 {\scriptsize[0.00,0.00]} & 0.00 {\scriptsize[0.00,0.00]} \\
\texttt{pile\_freelaw} & 0.49 {\scriptsize[0.41,0.57]} & 0.51 {\scriptsize[0.43,0.60]} & 0.60 {\scriptsize[0.54,0.65]} & 0.42 {\scriptsize[0.26,0.63]} & 0.38 {\scriptsize[0.00,0.88]} \\
\texttt{pile\_dm\_mathematics} & 0.43 {\scriptsize[0.37,0.49]} & 0.38 {\scriptsize[0.29,0.48]} & 0.45 {\scriptsize[0.39,0.51]} & 1.00 {\scriptsize[0.00,1.00]} & 0.73 {\scriptsize[0.44,1.00]} \\
\texttt{pile\_pubmed\_central} & 0.54 {\scriptsize[0.46,0.61]} & 0.44 {\scriptsize[0.33,0.57]} & 0.60 {\scriptsize[0.53,0.67]} & 0.40 {\scriptsize[0.28,0.53]} & 0.95 {\scriptsize[0.82,1.00]} \\
\texttt{pile\_wikipedia} & 0.49 {\scriptsize[0.42,0.57]} & 0.42 {\scriptsize[0.35,0.50]} & 0.58 {\scriptsize[0.52,0.64]} & 0.11 {\scriptsize[0.06,0.18]} & 0.86 {\scriptsize[0.71,1.00]} \\
\texttt{math500\_2021} & 0.42 {\scriptsize[0.34,0.51]} & 0.40 {\scriptsize[0.29,0.50]} & 0.52 {\scriptsize[0.46,0.59]} & 0.33 {\scriptsize[0.00,1.00]} & 1.00 {\scriptsize[0.00,1.00]} \\
\texttt{humaneval\_2021} & 0.24 {\scriptsize[0.12,0.36]} & 0.49 {\scriptsize[0.36,0.61]} & 0.38 {\scriptsize[0.31,0.45]} & 0.81 {\scriptsize[0.60,0.98]} & 0.08 {\scriptsize[0.00,0.50]} \\
\texttt{gsm8k\_2021} & 0.16 {\scriptsize[0.11,0.20]} & 0.27 {\scriptsize[0.22,0.32]} & 0.49 {\scriptsize[0.43,0.56]} & 0.01 {\scriptsize[0.00,0.03]} & 0.16 {\scriptsize[0.07,0.28]} \\
\texttt{ifeval\_2023} & 0.18 {\scriptsize[0.14,0.22]} & 0.22 {\scriptsize[0.16,0.27]} & 0.32 {\scriptsize[0.25,0.40]} & 0.33 {\scriptsize[0.00,1.00]} & 0.33 {\scriptsize[0.00,1.00]} \\
\texttt{gpqa\_diamond} & 0.20 {\scriptsize[0.15,0.25]} & 0.39 {\scriptsize[0.32,0.45]} & 0.57 {\scriptsize[0.51,0.63]} & n/a & n/a \\
\texttt{aime2026} & 0.30 {\scriptsize[0.18,0.41]} & 0.52 {\scriptsize[0.36,0.67]} & 0.50 {\scriptsize[0.37,0.61]} & n/a & n/a \\
\texttt{reddit\_2026} & 0.46 {\scriptsize[0.39,0.53]} & 0.48 {\scriptsize[0.40,0.56]} & 0.48 {\scriptsize[0.39,0.58]} & 0.20 {\scriptsize[0.00,0.55]} & 0.00 {\scriptsize[0.00,0.00]} \\
\bottomrule
\end{tabular}
}
\end{table}

\begin{table}[p]
\centering
\small
\caption{AUC scores (95\% CIs) separating seen and unseen corpora per model and method. \textbf{Pooled over documents}.}
\label{tab:ext-auc}
\begin{tabular*}{\linewidth}{@{\extracolsep{\fill}}lcccc@{}}
\toprule
Model & Carlini & SA-MIA & SimMIA & LCS (Ours) \\
\midrule
GPT-Neo 125M & 0.526 {\footnotesize[0.504,0.546]} & 0.578 {\footnotesize[0.553,0.601]} & 0.577 {\footnotesize[0.556,0.598]} & \textbf{0.822} {\footnotesize[0.807,0.838]} \\
GPT-Neo 1.3B & 0.536 {\footnotesize[0.514,0.558]} & 0.578 {\footnotesize[0.553,0.600]} & 0.609 {\footnotesize[0.587,0.631]} & \textbf{0.808} {\footnotesize[0.792,0.821]} \\
GPT-Neo 2.7B & 0.540 {\footnotesize[0.516,0.562]} & 0.565 {\footnotesize[0.542,0.588]} & 0.606 {\footnotesize[0.585,0.629]} & \textbf{0.812} {\footnotesize[0.798,0.825]} \\
Pythia 410M & 0.557 {\footnotesize[0.533,0.580]} & 0.565 {\footnotesize[0.541,0.588]} & 0.607 {\footnotesize[0.585,0.630]} & \textbf{0.811} {\footnotesize[0.799,0.826]} \\
Pythia 1.4B & 0.562 {\footnotesize[0.541,0.583]} & 0.569 {\footnotesize[0.546,0.591]} & 0.613 {\footnotesize[0.592,0.635]} & \textbf{0.816} {\footnotesize[0.801,0.829]} \\
Pythia 12B & 0.558 {\footnotesize[0.536,0.582]} & 0.559 {\footnotesize[0.535,0.582]} & 0.583 {\footnotesize[0.561,0.604]} & \textbf{0.796} {\footnotesize[0.780,0.810]} \\
RWKV-4 430M & 0.559 {\footnotesize[0.537,0.580]} & 0.572 {\footnotesize[0.549,0.595]} & 0.544 {\footnotesize[0.502,0.586]} & \textbf{0.785} {\footnotesize[0.770,0.799]} \\
RWKV-4 3B & 0.562 {\footnotesize[0.537,0.584]} & 0.565 {\footnotesize[0.543,0.589]} & 0.600 {\footnotesize[0.558,0.642]} & \textbf{0.792} {\footnotesize[0.777,0.808]} \\
RWKV-4 14B & 0.572 {\footnotesize[0.550,0.596]} & 0.557 {\footnotesize[0.533,0.581]} & 0.474 {\footnotesize[0.429,0.516]} & \textbf{0.724} {\footnotesize[0.709,0.741]} \\
OLMo 1B & 0.487 {\footnotesize[0.459,0.517]} & 0.459 {\footnotesize[0.427,0.491]} & 0.491 {\footnotesize[0.463,0.519]} & \textbf{0.696} {\footnotesize[0.671,0.721]} \\
OLMo 7B & 0.504 {\footnotesize[0.473,0.535]} & 0.452 {\footnotesize[0.419,0.484]} & 0.435 {\footnotesize[0.406,0.463]} & \textbf{0.717} {\footnotesize[0.696,0.743]} \\
\midrule
\textit{Mean} & 0.542 & 0.547 & 0.558 & \textbf{0.780} \\
\bottomrule
\end{tabular*}

\vspace{2em}

\caption{TPR@5\%FPR (95\% CIs) across models. Baselines use percentile bootstraps \citep{efron1979bootstrap}; LCS uses Clopper--Pearson bounds \citep{clopper1934use} on pooled counts. \textbf{Pooled over documents}.}
\label{tab:ext-tpr}
\begin{tabular*}{\linewidth}{@{\extracolsep{\fill}}lcccc@{}}
\toprule
Model & Carlini & SA-MIA & SimMIA & LCS (Ours) \\
\midrule
GPT-Neo 125M & 0.089 {\footnotesize[0.063,0.114]} & 0.168 {\footnotesize[0.147,0.197]} & 0.115 {\footnotesize[0.097,0.139]} & \textbf{0.302} {\footnotesize[0.276,0.328]} \\
GPT-Neo 1.3B & 0.097 {\footnotesize[0.076,0.125]} & \textbf{0.187} {\footnotesize[0.158,0.212]} & 0.139 {\footnotesize[0.113,0.169]} & 0.174 {\footnotesize[0.153,0.196]} \\
GPT-Neo 2.7B & 0.092 {\footnotesize[0.077,0.125]} & 0.172 {\footnotesize[0.147,0.210]} & 0.139 {\footnotesize[0.107,0.163]} & \textbf{0.174} {\footnotesize[0.153,0.196]} \\
Pythia 410M & 0.100 {\footnotesize[0.084,0.142]} & 0.166 {\footnotesize[0.138,0.191]} & 0.106 {\footnotesize[0.081,0.138]} & \textbf{0.289} {\footnotesize[0.264,0.315]} \\
Pythia 1.4B & 0.117 {\footnotesize[0.096,0.134]} & \textbf{0.181} {\footnotesize[0.152,0.203]} & 0.105 {\footnotesize[0.083,0.126]} & 0.158 {\footnotesize[0.138,0.179]} \\
Pythia 12B & 0.127 {\footnotesize[0.106,0.153]} & \textbf{0.193} {\footnotesize[0.158,0.222]} & 0.111 {\footnotesize[0.086,0.135]} & 0.148 {\footnotesize[0.129,0.169]} \\
RWKV-4 430M & 0.106 {\footnotesize[0.089,0.145]} & 0.186 {\footnotesize[0.160,0.220]} & 0.098 {\footnotesize[0.052,0.138]} & \textbf{0.213} {\footnotesize[0.190,0.236]} \\
RWKV-4 3B & 0.120 {\footnotesize[0.109,0.151]} & 0.197 {\footnotesize[0.171,0.218]} & 0.095 {\footnotesize[0.012,0.129]} & \textbf{0.214} {\footnotesize[0.192,0.238]} \\
RWKV-4 14B & 0.143 {\footnotesize[0.124,0.162]} & \textbf{0.178} {\footnotesize[0.159,0.199]} & 0.015 {\footnotesize[0.003,0.034]} & 0.051 {\footnotesize[0.040,0.065]} \\
OLMo 1B & 0.066 {\footnotesize[0.048,0.087]} & 0.069 {\footnotesize[0.046,0.092]} & 0.017 {\footnotesize[0.007,0.030]} & \textbf{0.098} {\footnotesize[0.076,0.125]} \\
OLMo 7B & \textbf{0.087} {\footnotesize[0.069,0.114]} & 0.067 {\footnotesize[0.048,0.091]} & 0.008 {\footnotesize[0.002,0.017]} & 0.085 {\footnotesize[0.064,0.110]} \\
\midrule
\textit{Mean} & 0.104 & 0.160 & 0.086 & \textbf{0.173} \\
\bottomrule
\end{tabular*}
\end{table}

\begin{table}[p]
\centering
\small
\caption{Dataset-level AUC (95\% CIs) separating seen and unseen corpora per model and method. \textbf{Pooled over datasets}.}
\label{tab:ext-auc-ds}
\begin{tabular*}{\linewidth}{@{\extracolsep{\fill}}lcccc@{}}
\toprule
Model & Carlini & SA-MIA & SimMIA & LCS (Ours) \\
\midrule
GPT-Neo 125M & 0.726 {\footnotesize[0.514,0.899]} & 0.558 {\footnotesize[0.327,0.774]} & 0.729 {\footnotesize[0.534,0.896]} & \textbf{0.977} {\footnotesize[0.910,1.000]} \\
GPT-Neo 1.3B & 0.776 {\footnotesize[0.577,0.928]} & 0.596 {\footnotesize[0.351,0.803]} & 0.801 {\footnotesize[0.620,0.937]} & \textbf{0.964} {\footnotesize[0.864,1.000]} \\
GPT-Neo 2.7B & 0.764 {\footnotesize[0.562,0.923]} & 0.572 {\footnotesize[0.327,0.788]} & 0.787 {\footnotesize[0.593,0.932]} & \textbf{0.977} {\footnotesize[0.910,1.000]} \\
Pythia 410M & 0.798 {\footnotesize[0.591,0.952]} & 0.582 {\footnotesize[0.351,0.798]} & 0.860 {\footnotesize[0.706,0.977]} & \textbf{0.968} {\footnotesize[0.891,1.000]} \\
Pythia 1.4B & 0.779 {\footnotesize[0.572,0.942]} & 0.601 {\footnotesize[0.361,0.817]} & 0.882 {\footnotesize[0.747,0.986]} & \textbf{0.968} {\footnotesize[0.891,1.000]} \\
Pythia 12B & 0.740 {\footnotesize[0.529,0.913]} & 0.548 {\footnotesize[0.303,0.774]} & 0.783 {\footnotesize[0.597,0.928]} & \textbf{0.950} {\footnotesize[0.824,1.000]} \\
RWKV-4 430M & 0.769 {\footnotesize[0.562,0.923]} & 0.596 {\footnotesize[0.356,0.808]} & 0.593 {\footnotesize[0.394,0.783]} & \textbf{0.973} {\footnotesize[0.891,1.000]} \\
RWKV-4 3B & 0.827 {\footnotesize[0.654,0.962]} & 0.591 {\footnotesize[0.356,0.808]} & 0.710 {\footnotesize[0.507,0.891]} & \textbf{0.959} {\footnotesize[0.864,1.000]} \\
RWKV-4 14B & 0.808 {\footnotesize[0.620,0.947]} & 0.591 {\footnotesize[0.351,0.808]} & 0.403 {\footnotesize[0.204,0.615]} & \textbf{0.968} {\footnotesize[0.878,1.000]} \\
OLMo 1B & 0.528 {\footnotesize[0.208,0.819]} & 0.472 {\footnotesize[0.153,0.792]} & 0.526 {\footnotesize[0.244,0.821]} & \textbf{0.859} {\footnotesize[0.538,1.000]} \\
OLMo 7B & 0.542 {\footnotesize[0.250,0.806]} & 0.472 {\footnotesize[0.167,0.792]} & 0.346 {\footnotesize[0.115,0.615]} & \textbf{0.910} {\footnotesize[0.692,1.000]} \\
\midrule
\textit{Mean} & 0.732 & 0.562 & 0.674 & \textbf{0.952} \\
\bottomrule
\end{tabular*}

\vspace{2em}

\caption{Dataset-level TPR@5\%FPR (95\% CIs) per model and method. \textbf{Pooled over datasets}.}
\label{tab:ext-tpr-ds}
\begin{tabular*}{\linewidth}{@{\extracolsep{\fill}}lcccc@{}}
\toprule
Model & Carlini & SA-MIA & SimMIA & LCS (Ours) \\
\midrule
GPT-Neo 125M & 0.154 {\footnotesize[0.000,0.538]} & 0.308 {\footnotesize[0.077,0.538]} & 0.077 {\footnotesize[0.000,0.538]} & \textbf{0.923} {\footnotesize[0.769,1.000]} \\
GPT-Neo 1.3B & 0.154 {\footnotesize[0.000,0.692]} & 0.385 {\footnotesize[0.154,0.692]} & 0.154 {\footnotesize[0.000,0.769]} & \textbf{0.923} {\footnotesize[0.769,1.000]} \\
GPT-Neo 2.7B & 0.154 {\footnotesize[0.000,0.692]} & 0.385 {\footnotesize[0.154,0.692]} & 0.077 {\footnotesize[0.000,0.615]} & \textbf{0.923} {\footnotesize[0.769,1.000]} \\
Pythia 410M & 0.615 {\footnotesize[0.308,0.846]} & 0.308 {\footnotesize[0.077,0.538]} & 0.154 {\footnotesize[0.000,0.846]} & \textbf{0.846} {\footnotesize[0.692,1.000]} \\
Pythia 1.4B & 0.538 {\footnotesize[0.308,0.846]} & 0.385 {\footnotesize[0.154,0.692]} & 0.231 {\footnotesize[0.077,0.846]} & \textbf{0.923} {\footnotesize[0.769,1.000]} \\
Pythia 12B & 0.538 {\footnotesize[0.308,0.846]} & 0.308 {\footnotesize[0.077,0.615]} & 0.000 {\footnotesize[0.000,0.538]} & \textbf{0.923} {\footnotesize[0.769,1.000]} \\
RWKV-4 430M & 0.538 {\footnotesize[0.308,0.846]} & 0.308 {\footnotesize[0.077,0.615]} & 0.000 {\footnotesize[0.000,0.308]} & \textbf{0.923} {\footnotesize[0.769,1.000]} \\
RWKV-4 3B & 0.462 {\footnotesize[0.231,0.846]} & 0.385 {\footnotesize[0.154,0.692]} & 0.000 {\footnotesize[0.000,0.692]} & \textbf{0.846} {\footnotesize[0.692,1.000]} \\
RWKV-4 14B & 0.385 {\footnotesize[0.154,0.846]} & 0.385 {\footnotesize[0.154,0.692]} & 0.000 {\footnotesize[0.000,0.000]} & \textbf{0.923} {\footnotesize[0.769,1.000]} \\
OLMo 1B & 0.167 {\footnotesize[0.000,0.500]} & 0.167 {\footnotesize[0.000,0.667]} & 0.000 {\footnotesize[0.000,0.333]} & \textbf{0.833} {\footnotesize[0.500,1.000]} \\
OLMo 7B & 0.167 {\footnotesize[0.000,0.500]} & 0.167 {\footnotesize[0.000,0.500]} & 0.000 {\footnotesize[0.000,0.000]} & \textbf{0.833} {\footnotesize[0.500,1.000]} \\
\midrule
\textit{Mean} & 0.352 & 0.317 & 0.063 & \textbf{0.893} \\
\bottomrule
\end{tabular*}
\end{table}

\clearpage
\section{Uncovering the ``Tells'': Extended Qualitative Examples}
\label{app:qualitative-examples}

We visualize context-induced token shifts ($\delta_j$) on Pythia-410M \citep{biderman2023pythia}, comparing a memorized pre-training corpus (\texttt{pile\_pubmed\_abstracts}) \citep{gao2020pile} with a held-out one (\texttt{bbc\_news\_2025}). Green ($\delta_j = 1$) and red ($\delta_j = -1$) highlights mark positions where the context ($K=3$) corrects or disrupts the prediction, respectively.

On seen documents (Figure~\ref{fig:app-seen-tells}), the high density of red tokens illustrates that context interferes with the exact-match priors of memorized text \citep{xie2024recall, codec}.

\begin{figure}[H]
  \centering
  \fbox{\parbox{0.98\linewidth}{\footnotesize
    \providecommand{\dahelp}[2]{\colorbox{green!#2!white}{\strut #1}}
    \providecommand{\dahurt}[2]{\colorbox{red!#2!white}{\strut #1}}
    \providecommand{\daidle}[1]{#1}
    \providecommand{\dadocsepskip}{3pt}
    \providecommand{\dadocsep}{\par\addvspace{\dadocsepskip}\noindent
      \hbox to\linewidth{\leaders\hbox to .7em{\hss.\hss}\hfil}%
      \par\addvspace{\dadocsepskip}\noindent\ignorespaces}
    
    \textbf{Seen Documents:} \texttt{pile\_pubmed\_abstracts} \hfill \textit{Context hurts: 11 helped (green), 41 hurt (red)}
    \vspace{2pt}
    
    {\setlength{\fboxsep}{1pt}\sloppy\par\noindent
    \daidle{A} \daidle{monoclonal} \daidle{antibody}\daidle{,} \daidle{MN}\daidle{13}\daidle{,} \daidle{that} \dahurt{recognizes}{70} \daidle{specifically} \daidle{a} \daidle{novel} \daidle{substance} \daidle{between} \daidle{the} \daidle{post}\daidle{ac}\dahurt{ros}{70}\daidle{omal} \daidle{sheath} \daidle{and} \daidle{the} \daidle{over}\daidle{lying} \daidle{plasma} \daidle{membrane} \daidle{in} \daidle{the} \daidle{mammalian} \dahurt{sperm}{70} \daidle{head}\daidle{.} \daidle{Mon}\daidle{oclonal} \daidle{antibody} \daidle{MN}\daidle{13} \daidle{raised} \dahurt{against}{70} \daidle{mouse} \daidle{spermat}\daidle{oz}\daidle{oa} \daidle{specifically} \dahelp{recognizes}{70} \daidle{the} \dahurt{post}{70}\daidle{ac}\daidle{ros}\daidle{omal} \daidle{region} \daidle{of} \daidle{the} \dahurt{sperm}{70} \daidle{head} \daidle{in} \daidle{several} \daidle{mammalian} \daidle{species}\daidle{.} \daidle{Coll}\daidle{oidal} \daidle{gold}\daidle{-}\daidle{immun}\daidle{oelect}\dahurt{ron}{70} \daidle{microscopy} \dahurt{of}{70} \daidle{dem}\daidle{em}\daidle{br}\daidle{an}\daidle{ated} \daidle{mouse} \daidle{spermat}\daidle{oz}\daidle{oa} \daidle{indicated} \daidle{that} \daidle{the} \daidle{antigen} \daidle{is} \daidle{associated} \daidle{with} \daidle{the} \daidle{outer} \daidle{layer} \daidle{of} \daidle{the} \daidle{periodic} \daidle{sub}\daidle{structure} \daidle{apparently} \daidle{linking} \daidle{the} \daidle{post}\daidle{ac}\daidle{ros}\daidle{omal} \dahelp{sheath}{70} \daidle{to} \daidle{the} \daidle{over}\daidle{lying} \daidle{plasma} \daidle{membrane}\daidle{.} \dahurt{The}{70} \dahurt{antigen}{70}\par}

    \dadocsep
    
    {\setlength{\fboxsep}{1pt}\sloppy\par\noindent
    \daidle{Gen}\daidle{etic} \dahurt{and}{70} \dahurt{clinical}{70} \dahurt{features}{70} \daidle{of} \daidle{p}\daidle{450} \daidle{oxid}\daidle{ore}\daidle{duct}\daidle{ase} \dahurt{deficiency}{70}\daidle{.} \daidle{P}\dahurt{450}{70} \daidle{oxid}\daidle{ore}\daidle{duct}\daidle{ase} \daidle{(}\daidle{POR}\daidle{)} \daidle{deficiency} \daidle{is} \daidle{an} \daidle{autosomal} \daidle{recess}\daidle{ive} \daidle{disorder} \daidle{of} \daidle{steroid}\dahelp{ogenesis}{70} \daidle{with} \daidle{multiple} \daidle{clinical} \daidle{manifestations}\daidle{.} \daidle{P}\daidle{OR} \daidle{is} \daidle{the} \daidle{electron} \daidle{donor} \daidle{for} \daidle{all} \daidle{micro}\daidle{s}\daidle{omal} \daidle{P}\daidle{450} \dahurt{enzymes}{70}\dahurt{,}{70} \dahelp{including}{70} \dahurt{the}{70} \daidle{three} \daidle{steroid}\daidle{ogenic} \daidle{enzymes} \daidle{P}\daidle{450}\dahelp{c}{70}\daidle{17} \daidle{(}\daidle{17}\daidle{alpha}\daidle{-}\daidle{hydrox}\daidle{ylase}\daidle{/}\daidle{17}\daidle{,}\daidle{20}\daidle{-}\daidle{ly}\daidle{ase}\daidle{),} \daidle{P}\daidle{450}\daidle{c}\daidle{21} \daidle{(}\daidle{21}\dahurt{-}{70}\daidle{hydrox}\daidle{ylase}\daidle{),} \daidle{and} \daidle{P}\daidle{450}\daidle{aro} \daidle{(}\daidle{ar}\daidle{omat}\daidle{ase}\daidle{).} \daidle{Since} \daidle{the} \daidle{first} \daidle{description} \daidle{of} \daidle{P}\daidle{OR} \daidle{mutations} \daidle{in} \daidle{2004}\daidle{,} \daidle{about} \daidle{50} \daidle{patients}\par}

    \dadocsep
    
    {\setlength{\fboxsep}{1pt}\sloppy\par\noindent
    \daidle{M}\daidle{orph}\daidle{ometric}\dahurt{s}{70} \dahelp{of}{70} \daidle{a} \daidle{ne}\daidle{ot}\dahurt{ropical}{70} \daidle{sand}\dahurt{fly}{70} \daidle{sub}\dahurt{species}{70}\daidle{,} \daidle{L}\daidle{utz}\daidle{omy}\daidle{ia} \daidle{car}\daidle{rera}\daidle{i} \daidle{th}\daidle{ula}\daidle{.} \daidle{In} \daidle{spite} \daidle{of} \daidle{obvious} \daidle{and} \daidle{frequently} \daidle{observed} \daidle{morphological} \daidle{variation} \dahelp{in}{70} \daidle{geographic} \daidle{populations} \daidle{of} \daidle{sand} \daidle{flies}\daidle{,} \daidle{tax}\daidle{onom}\dahurt{ists}{70} \daidle{are} \daidle{reluctant} \daidle{to} \daidle{create} \daidle{sub}\daidle{species}\daidle{.} \daidle{In} \daidle{doing} \daidle{so} \daidle{they} \daidle{avoid} \daidle{confusion} \daidle{caused} \daidle{by} \daidle{the} \daidle{accumulation} \daidle{of} \daidle{new} \daidle{taxa}\dahurt{,}{70} \daidle{but} \daidle{they} \daidle{also} \daidle{obscure} \daidle{natural} \daidle{processes}\daidle{,} \daidle{some} \daidle{of} \daidle{them} \daidle{related} \daidle{to} \daidle{probable} \daidle{species} \daidle{emergence}\daidle{.} \daidle{We} \daidle{investigated} \daidle{here} \daidle{how} \daidle{far} \daidle{the} \daidle{partitioning} \daidle{of} \daidle{size}\daidle{,} \daidle{shape} \dahurt{and}{70} \daidle{form} \daidle{could} \daidle{be} \daidle{a} \daidle{relevant} \daidle{technique} \daidle{for} \daidle{exploring} \daidle{early} \daidle{evolutionary} \dahurt{processes}{70}\daidle{.} \daidle{We} \daidle{used} \daidle{a} \daidle{typical} \daidle{case} \daidle{of}\par}

    \dadocsep
    
    {\setlength{\fboxsep}{1pt}\sloppy\par\noindent
    \daidle{Effect}\daidle{ively} \daidle{addressing} \daidle{the} \daidle{mid}\daidle{-} \daidle{and} \daidle{long}\daidle{-}\daidle{term} \daidle{needs} \daidle{of} \daidle{young} \dahurt{people}{70} \daidle{affected} \daidle{by} \daidle{the} \daidle{t}\daidle{sunami} \daidle{in} \daidle{Ace}\daidle{h}\daidle{:} \daidle{an} \daidle{on}\daidle{-}\daidle{site} \daidle{assessment}\daidle{.} \daidle{Within} \daidle{two} \daidle{months} \daidle{of} \daidle{the} \daidle{Asian} \daidle{t}\daidle{sunami}\daidle{,} \daidle{a} \dahurt{team}{70} \daidle{of} \daidle{four} \daidle{individuals} \daidle{conducted} \daidle{an} \daidle{assessment} \daidle{on} \daidle{the} \daidle{post}\daidle{-}\dahelp{dis}{70}\daidle{aster} \daidle{needs} \daidle{of} \dahelp{young}{70} \daidle{people} \dahurt{in}{70} \daidle{Ace}\daidle{h} \daidle{Province}\dahurt{.}{70} \daidle{In} \daidle{addition} \daidle{to} \daidle{assessing} \daidle{current} \daidle{needs}\daidle{,} \daidle{the} \daidle{team} \daidle{examined} \daidle{the} \daidle{extent} \dahurt{to}{70} \daidle{which} \daidle{young} \daidle{people} \daidle{(}\daidle{aged} \daidle{14}\daidle{-}\daidle{24}\daidle{)} \dahurt{were}{70} \daidle{involved} \daidle{in} \daidle{the} \daidle{planning} \daidle{and} \daidle{implementation} \daidle{of} \daidle{ongoing} \daidle{rebuilding} \daidle{and} \daidle{relief} \daidle{efforts}\daidle{.} \daidle{Finally}\daidle{,} \daidle{the} \daidle{team} \daidle{assessed} \daidle{the} \daidle{degree} \dahurt{to}{70} \daidle{which} \daidle{young} \daidle{people} \daidle{could} \daidle{be}\par}

    \dadocsep
    
    {\setlength{\fboxsep}{1pt}\sloppy\par\noindent
    \daidle{Effect} \daidle{of} \daidle{hal}\dahelp{oper}{70}\daidle{id}\daidle{ol} \daidle{and} \daidle{ris}\daidle{per}\daidle{idone} \daidle{on} \daidle{amyloid} \daidle{precursor} \daidle{protein} \daidle{levels} \daidle{in} \daidle{vivo}\dahurt{.}{70} \dahurt{The}{70} \daidle{neuro}\daidle{toxic} \daidle{beta}\daidle{-}\daidle{amyl}\daidle{oid} \daidle{peptide} \daidle{of} \dahurt{Alzheimer}{70}\daidle{'s} \daidle{disease} \daidle{is} \daidle{formed} \daidle{from} \daidle{the} \dahurt{amyloid}{70} \daidle{precursor} \daidle{protein} \daidle{(}\daidle{APP}\daidle{),} \daidle{which} \daidle{is} \daidle{a} \daidle{member} \daidle{of} \daidle{an} \daidle{evolution}\daidle{arily} \daidle{highly} \daidle{conserved} \daidle{gene} \daidle{family} \daidle{with} \daidle{significant} \daidle{functional} \daidle{importance}\daidle{.} \daidle{Because} \daidle{behavioral} \daidle{and} \daidle{psychiatric} \daidle{symptoms} \daidle{treated} \daidle{with} \daidle{antipsych}\daidle{otics} \daidle{may} \daidle{influence} \dahurt{the}{70} \daidle{course} \daidle{of} \daidle{the} \daidle{disease}\daidle{,} \dahelp{we}{70} \daidle{have} \dahurt{investigated}{70} \daidle{traditional} \daidle{and} \daidle{atypical} \daidle{antipsych}\daidle{otic} \dahurt{drugs}{70}\daidle{,} \daidle{administered} \daidle{through} \daidle{the} \daidle{intra}\daidle{peritoneal} \daidle{route}\daidle{,} \daidle{for} \daidle{their} \daidle{effects} \daidle{on} \daidle{rat} \daidle{cortical} \daidle{APP}\daidle{.} \daidle{Western}\daidle{-}\daidle{immun}\daidle{oblot}\daidle{ting} \daidle{was} \daidle{utilized} \daidle{for} \daidle{semi}\daidle{-}\daidle{quant}\par}
  }}
  \caption{Positions where adding context changes the next-token prediction for \textbf{seen} documents. Red highlights ($\delta_j = -1$) heavily dominate, showing that context interferes with the model's exact-match verbatim memorization.}
  \label{fig:app-seen-tells}
\end{figure}

\newpage

Conversely, Figure~\ref{fig:app-unseen-tells} demonstrates the model's behavior on unseen documents. Because these recent news articles post-date the pre-training cutoff, the model has no exact-match priors to disrupt. Instead, it relies heavily on in-context learning \citep{brown2020language}. The provided prefix serves as a beneficial semantic anchor, leading to predominantly green shifts as the context actively aids the next-token predictions rather than competing against memorized weights.

\begin{figure}[H]
  \centering
  \fbox{\parbox{0.98\linewidth}{\footnotesize
    \providecommand{\dahelp}[2]{\colorbox{green!#2!white}{\strut #1}}
    \providecommand{\dahurt}[2]{\colorbox{red!#2!white}{\strut #1}}
    \providecommand{\daidle}[1]{#1}
    \providecommand{\dadocsepskip}{3pt}
    \providecommand{\dadocsep}{\par\addvspace{\dadocsepskip}\noindent
      \hbox to\linewidth{\leaders\hbox to .7em{\hss.\hss}\hfil}%
      \par\addvspace{\dadocsepskip}\noindent\ignorespaces}
    
    \textbf{Unseen Documents:} \texttt{bbc\_news\_2025} \hfill \textit{Context helps: 51 helped (green), 10 hurt (red)}
    \vspace{2pt}
    
    {\setlength{\fboxsep}{1pt}\sloppy\par\noindent
    \daidle{William} \daidle{shocked} \dahelp{by}{70} \dahelp{death}{70} \daidle{of} \daidle{ex}\daidle{-}\dahelp{n}{70}\dahelp{anny}{70}\daidle{'s} \daidle{step}\daidle{son} \daidle{in} \dahelp{New}{70} \dahelp{Orleans}{70} \daidle{attack}\par\noindent \daidle{Edward} \dahelp{P}{70}\dahelp{ett}{70}\dahelp{ifer}{70} \daidle{has} \daidle{been} \daidle{identified} \daidle{by} \daidle{the} \daidle{Metropolitan} \daidle{Police} \daidle{as} \daidle{the} \daidle{British} \daidle{national} \daidle{who} \daidle{was} \daidle{killed} \daidle{in} \daidle{New} \daidle{Orleans} \dahelp{on}{70} \daidle{1} \dahelp{January}{70}\par\noindent \daidle{The} \daidle{Prince} \daidle{of} \daidle{Wales} \daidle{has} \daidle{said} \daidle{he} \daidle{and} \daidle{Catherine} \daidle{are} \dahelp{"}{70}\daidle{shock}\daidle{ed} \daidle{and} \daidle{sadd}\daidle{ened}\daidle{"} \daidle{by} \daidle{the} \daidle{death} \daidle{of} \daidle{his} \daidle{former} \daidle{n}\daidle{anny}\daidle{'s} \daidle{step}\daidle{son}\dahelp{,}{70} \daidle{Edward} \daidle{P}\daidle{ett}\daidle{ifer}\daidle{,} \daidle{who} \daidle{was} \daidle{killed} \daidle{in} \dahurt{the}{70} \daidle{New} \daidle{Orleans} \daidle{attack}\dahurt{.}{70} \dahelp{The}{70} \daidle{31}\daidle{-}\daidle{year}\daidle{-}\daidle{old} \dahurt{was}{70} \daidle{among} \daidle{14} \daidle{people} \daidle{killed} \daidle{in} \daidle{a} \daidle{vehicle} \dahelp{attack}{70} \daidle{on} \dahelp{New}{70} \daidle{Year}\daidle{'s} \dahelp{Day}{70}\dahurt{.}{70} \daidle{His} \daidle{step}\daidle{mother}\par}

    \dadocsep
    
    {\setlength{\fboxsep}{1pt}\sloppy\par\noindent
    \daidle{Far}\dahelp{age}{70} \daidle{distances} \dahelp{himself}{70} \dahelp{from}{70} \daidle{Musk} \daidle{on} \daidle{Robinson}\par\noindent \daidle{El}\daidle{on} \daidle{Musk} \daidle{(}\daidle{L}\daidle{)} \daidle{and} \dahelp{Nig}{70}\daidle{el} \daidle{Far}\daidle{age} \daidle{at} \daidle{Donald} \daidle{Trump}\daidle{'s} \daidle{Mar}\daidle{-}\daidle{A}\daidle{-}\daidle{L}\daidle{ago} \daidle{home} \daidle{in} \daidle{Florida} \daidle{last} \daidle{month}\par\noindent \daidle{Re}\dahelp{form}{70} \dahelp{UK}{70} \daidle{leader} \daidle{Nig}\daidle{el} \daidle{Far}\daidle{age} \dahelp{has}{70} \daidle{dist}\daidle{anced} \daidle{himself} \daidle{from} \daidle{El}\daidle{on} \daidle{Musk}\daidle{'s} \daidle{support} \daidle{for} \daidle{j}\daidle{ailed} \daidle{far}\daidle{-}\daidle{right} \daidle{activist} \daidle{Stephen} \daidle{Y}\daidle{ax}\daidle{ley}\daidle{-}\daidle{L}\daidle{enn}\daidle{on}\dahurt{,}{70} \daidle{also} \daidle{known} \daidle{as} \daidle{Tommy} \daidle{Robinson}\dahelp{.}{70} \dahelp{The}{70} \daidle{tech} \daidle{multi}\daidle{-}\daidle{billion}\daidle{aire} \daidle{added} \daidle{his} \dahelp{voice}{70} \daidle{to} \daidle{calls} \daidle{to} \daidle{release} \daidle{Y}\daidle{ax}\daidle{ley}\daidle{-}\daidle{L}\daidle{enn}\daidle{on}\daidle{,} \daidle{who} \daidle{was} \daidle{j}\daidle{ailed} \dahurt{in}{70} \daidle{October} \daidle{after} \daidle{admitting} \daidle{contempt} \daidle{of} \daidle{court} \daidle{by}\par}

    \dadocsep
    
    {\setlength{\fboxsep}{1pt}\sloppy\par\noindent
    \daidle{St}\daidle{orm}\daidle{zy} \daidle{banned} \dahelp{from}{70} \daidle{driving} \dahelp{after}{70} \daidle{mobile} \daidle{phone} \daidle{use}\par\noindent \dahelp{The}{70} \daidle{31}\daidle{-}\daidle{year}\daidle{-}\daidle{old} \daidle{rapper} \daidle{already} \daidle{had} \daidle{six} \daidle{points} \daidle{on} \daidle{his} \dahelp{licence}{70}\par\noindent \daidle{St}\daidle{orm}\daidle{zy} \daidle{has} \daidle{been} \daidle{banned} \daidle{from} \daidle{driving} \daidle{for} \daidle{nine} \daidle{months} \daidle{after} \daidle{an} \daidle{off}\daidle{-}\daidle{duty} \daidle{police} \daidle{officer} \daidle{caught} \daidle{him} \daidle{using} \daidle{a} \daidle{mobile} \daidle{phone} \daidle{behind} \dahelp{the}{70} \daidle{wheel} \daidle{of} \daidle{a} \daidle{Roll}\daidle{s}\daidle{-}\daidle{Roy}\daidle{ce} \daidle{in} \dahelp{London}{70}\daidle{.} \dahelp{The}{70} \daidle{31}\daidle{-}\daidle{year}\daidle{-}\daidle{old} \daidle{rapper}\dahelp{,}{70} \daidle{whose} \daidle{real} \daidle{name} \daidle{is} \daidle{Michael} \daidle{E}\daidle{ben}\daidle{azer} \daidle{Ow}\daidle{uo} \daidle{Junior}\daidle{,} \daidle{pleaded} \daidle{guilty} \daidle{via} \daidle{a} \daidle{letter} \daidle{to} \daidle{driving} \daidle{the} \dahurt{Roll}{70}\daidle{s}\daidle{-}\daidle{Roy}\daidle{ce} \daidle{W}\daidle{ra}\daidle{ith} \daidle{while} \daidle{using} \daidle{a} \daidle{device} \daidle{on} \daidle{Add}\daidle{ison} \daidle{Road}\daidle{,} \daidle{west}\par}

    \dadocsep
    
    {\setlength{\fboxsep}{1pt}\sloppy\par\noindent
    \daidle{French} \daidle{sk}\dahelp{ier}{70} \daidle{Cyp}\daidle{rien} \daidle{Sar}\daidle{raz}\daidle{in} \daidle{has} \daidle{left} \daidle{intensive} \dahelp{care}{70} \dahelp{after}{70} \daidle{successful} \daidle{surgery} \daidle{to} \daidle{drain} \daidle{a} \daidle{bleed} \daidle{on} \daidle{his} \dahelp{brain}{70} \daidle{following} \daidle{a} \dahelp{fall}{70} \daidle{in} \daidle{training}\daidle{.}\par\noindent \daidle{Sar}\daidle{raz}\daidle{in}\daidle{,} \daidle{30}\daidle{,} \daidle{was} \daidle{air}\daidle{lift}\daidle{ed} \daidle{to} \daidle{hospital} \daidle{after} \daidle{losing} \daidle{control} \daidle{while} \daidle{training} \daidle{for} \daidle{the} \daidle{World} \daidle{Cup} \daidle{downhill} \daidle{in} \daidle{B}\daidle{orm}\daidle{io}\daidle{,} \daidle{Italy}\daidle{,} \daidle{on} \daidle{Friday}\daidle{.} \daidle{He} \daidle{had} \daidle{surgery} \daidle{the} \daidle{same} \daidle{day} \dahelp{to}{70} \daidle{reduce} \daidle{an} \daidle{intracranial} \daidle{haemat}\daidle{oma}\dahurt{.}{70}\par\noindent \daidle{"}\daidle{His} \daidle{state} \dahelp{of}{70} \daidle{health} \daidle{is} \daidle{stable}\daidle{.} \daidle{He} \daidle{was} \daidle{discharged} \daidle{from} \daidle{intensive} \daidle{care} \daidle{and} \daidle{admitted} \daidle{to} \daidle{a} \daidle{care} \daidle{unit}\daidle{,"} \dahelp{the}{70} \daidle{French} \daidle{Ski} \daidle{Federation} \daidle{posted} \daidle{on} \daidle{X} \daidle{on} \daidle{Thursday}\daidle{.}\par\noindent \daidle{"}\par}

    \dadocsep
    
    {\setlength{\fboxsep}{1pt}\sloppy\par\noindent
    \daidle{The} \daidle{film} \daidle{W}\daidle{icked} \daidle{has} \daidle{been} \dahurt{a}{70} \daidle{huge} \dahelp{hit}{70} \daidle{with} \daidle{fans}\daidle{,} \daidle{becoming} \daidle{one} \daidle{of} \daidle{the} \daidle{highest}\dahelp{-}{70}\daidle{g}\daidle{ross}\daidle{ing} \daidle{movies} \daidle{of} \daidle{20}\daidle{24}\daidle{.}\par\noindent \daidle{Ada}\daidle{pt}\dahelp{ed}{70} \daidle{from} \daidle{the} \daidle{stage} \daidle{musical}\dahelp{,}{70} \daidle{it} \dahelp{tells}{70} \daidle{the} \daidle{origin} \daidle{story} \daidle{of} \daidle{El}\daidle{ph}\daidle{aba}\daidle{,} \daidle{better} \daidle{known} \daidle{as} \daidle{the} \daidle{W}\daidle{icked} \daidle{Witch} \daidle{of} \daidle{the} \daidle{West} \daidle{from} \daidle{The} \daidle{Wizard} \daidle{of} \daidle{Oz}\daidle{.}\par\noindent \daidle{El}\daidle{ph}\daidle{aba}\daidle{'s} \daidle{wig} \daidle{was} \daidle{bra}\daidle{ided} \daidle{in} \daidle{the} \daidle{capital} \daidle{by} \dahelp{a}{70} \daidle{woman} \daidle{who} \daidle{came} \daidle{to} \daidle{London} \daidle{as} \daidle{a} \daidle{Cong}\daidle{ol}\daidle{ese} \daidle{refugee} \daidle{and} \daidle{now} \daidle{runs} \daidle{a} \daidle{hair} \daidle{salon} \daidle{in} \daidle{Seven} \daidle{Sisters}\daidle{.}\par\noindent \daidle{This} \daidle{article} \dahelp{was}{70} \daidle{amended} \daidle{to} \daidle{make} \daidle{clear} \daidle{the} \daidle{wig} \daidle{was} \daidle{made}\par}
  }}
  \caption{Positions where adding context changes the next-token prediction for \textbf{unseen} documents. Green highlights ($\delta_j = 1$) dominate, demonstrating that semantic context actively aids the model through in-context learning.}
  \label{fig:app-unseen-tells}
\end{figure}
\end{document}